\documentclass{article}

\usepackage[final]{neurips_2022}

\usepackage[utf8]{inputenc} %
\usepackage[T1]{fontenc}    %
\usepackage{hyperref}       %
\usepackage{url}            %
\usepackage{booktabs}       %
\usepackage{amsfonts}       %
\usepackage{nicefrac}       %
\usepackage{microtype}      %
\usepackage{color}
\usepackage{amsmath}
\usepackage{amsthm}
\usepackage{graphicx}
\usepackage{subcaption}
\usepackage{comment}
\usepackage{tikz}
\usepackage{natbib}
\usepackage{caption}
\usepackage{multirow}
\usepackage{wrapfig}
\usepackage{algorithm, algpseudocode}%
\usepackage{amsfonts, amsmath}
\usepackage{enumitem}
\usepackage{subcaption}
\usepackage{soul}
\usepackage{bm}
\usepackage{wrapfig}
\usepackage{listings}
\usepackage[title]{appendix}
\usepackage[bottom]{footmisc}
\usepackage[makeroom]{cancel}
\usepackage[normalem]{ulem}
\setcitestyle{authoryear,open={(},close={)}}
\usepackage{array}
\usepackage{algorithm}
\usepackage{algpseudocode}
\newcolumntype{P}[1]{>{\centering\arraybackslash}p{#1}}
\usepackage{varwidth}
\usepackage{lipsum}
\usepackage{multicol}
\newsavebox{\algbox}
\algrenewcommand{\Return}{\State \textbf{return}~}
\usepackage{import}
\usetikzlibrary{bayesnet}
\usetikzlibrary{arrows}

\usepackage[utf8]{inputenc} % allow utf-8 input
\usepackage[T1]{fontenc}    % use 8-bit T1 fonts
\usepackage{hyperref}       % hyperlinks
\usepackage{url}            % simple URL typesetting
\usepackage{booktabs}       % professional-quality tables
\usepackage{amsfonts}       % blackboard math symbols
\usepackage{nicefrac}       % compact symbols for 1/2, etc.
\usepackage{microtype}      % microtypography
\usepackage{xcolor}         % colors

\usepackage{pifont}
\newcommand{\cmark}{ \textcolor{green!60!black}{\ding{51}} }
\newcommand{\xmark}{ \textcolor{red!60!black}{\ding{55}} }

\def\y{\mathbf{y}}

\def\dz{\dot{\mathbf{z}}}

\def\v{\mathbf{v}}

\def\bmu{\boldsymbol{\mu}}
\def\z{\mathbf{z}}
\def\bt{{\boldsymbol{\theta}}}
\def\0{\mathbf{0}}
\def\R{\mathbb{R}}
\def\D{\mathcal{D}}
\def\A{\mathcal{A}}

\def\bl{\boldsymbol{\lambda}}
\def\B{\mathcal{B}}

\def\N{\mathcal{N}}
\def\L{\mathcal{L}}

\def\E{\mathbb{E}}

\def\atC{\texttt{C}}
\def\atH{\texttt{H}}
\def\atN{\texttt{N}}
\def\atO{\texttt{O}}
\def\atP{\texttt{P}}
\def\atF{\texttt{F}}
\def\atS{\texttt{S}}
\def\atMod{\texttt{ModFlow}}

\DeclareMathOperator*{\Cat}{Cat}

\DeclareMathOperator*{\tr}{tr}
\DeclareMathOperator*{\kl}{KL}
\DeclareRobustCommand{\rchi}{{\mathpalette\irchi\relax}}
\newcommand{\irchi}[2]{\raisebox{\depth}{$#1\chi$}}
\title{Modular Flows: Differential Molecular Generation}

\author{
  Yogesh Verma, Samuel Kaski, Markus Heinonen  \\
  Aalto University\\
  \texttt{\{yogesh.verma, samuel.kaski, markus.heinonen\}@aalto.fi} \\
   \And
   Vikas Garg \\
  YaiYai Ltd and Aalto University\\
  \texttt{vgarg@csail.mit.edu; vikas@yaiyai.fi} \\
}

\begin{document}

\maketitle
% \color{blue} Vikas: I will add some comments/suggestions in blue. 
% \begin{itemize}
%     \item Changing name to Modulow for now 
% \end{itemize}
\color{black}

\begin{abstract}
Generating new molecules is fundamental to advancing critical applications such as drug discovery and material synthesis. Flows can generate molecules effectively by inverting the encoding process, however, existing flow models either require artifactual dequantization or specific node/edge orderings, lack desiderata such as permutation invariance, or induce discrepancy between the encoding and the decoding steps that necessitates {\em post hoc} validity correction. We circumvent these issues with novel continuous normalizing E(3)-equivariant flows, based on a system of node ODEs coupled as a graph PDE, that repeatedly reconcile locally toward globally aligned densities. Our models can be cast as message passing temporal networks, and result in superlative performance on the tasks of density estimation and  molecular generation. In particular, our generated samples achieve state of the art on both the standard QM9 and ZINC250K benchmarks. 
\end{abstract}

\iffalse
Generating valid molecular structures is currently a fundamental problem for drug discovery which has been attracting growing attention.  Inspired by the progress in generative models, we propose Modular Flows (\atMod), a novel latent variable model for molecular graph generation based on continous normalizing flows. \atMod~  incorporates the E(3) Equivariant coupled ODE dynamical system over graphs into a generative model by using normalizing flows for molecular generation tasks. In our method, each node of the graph is coupled via an ODE system to its neighbouring nodes, thus constraining the system to flow in a coupled manner. Thus, attacking the global validity of molecule via accurate local densities. We demonstrate the applicability of our model on multiple tasks ranging from molecular generation to density estimation.  Experimentally, the proposed method significantly outperforms previous molecular generative methods regarding the quality of generated samples.    
\fi

\section{Introduction}

%Generative models have recently propelled several scientific disciplines enabling notable advances such as in caption-conditional image synthesis \citep{Ramesh2022}, protein design \citep{ingraham2019}, and single-cell transcriptomics \citep{lopez2018generative}, etc.  

\begin{wrapfigure}[20]{r}{0.43\textwidth}
    \vspace{-10pt}
    \hspace{-9pt}
    \includegraphics[scale=0.25, trim={0 0.1cm 0 0}, clip]{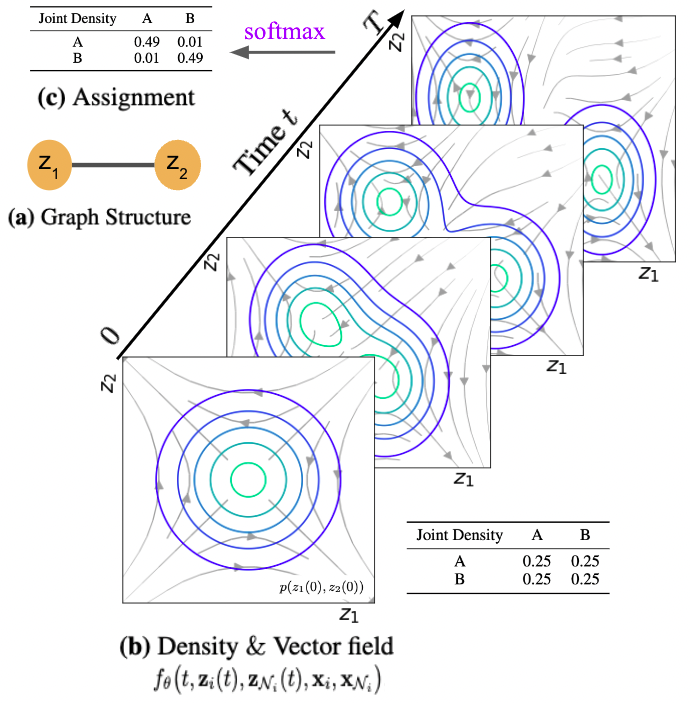}
    \caption{A toy illustration of \atMod~  in action with a two-node graph. The two local flows - $\z_1$ and $\z_2$ - co-evolve toward a more complex joint density, both driven by the same differential $f$.}
    \label{fig:placeholder}
\end{wrapfigure}
Generative models have rapidly become ubiquitous in machine learning with advances from image synthesis \citep{Ramesh2022} to protein design \citep{ingraham2019}. %nd molecular generation \citep{STOKES2020688}.
Molecular generation \citep{STOKES2020688} has also received significant attention owing to its promise for discovering new drugs and materials. Searching for valid molecules in prohibitively large discrete spaces is, however, challenging: estimates for drug-like structures range between $10^{23}$ and $10^{60}$ but only a tiny fraction - on the order of $10^8$ - has been synthesized  \citep{polishchuk2013estimation, Merz2020}. Thus, learning representations that exploit appropriate molecular inductive biases (e.g., spatial correlations) becomes crucial.

\begin{table}[!b]
    \caption{A comparison of generative modeling approaches for molecules.}
    \label{tab:relatedworks}
    \centering
%    \resizebox{\textwidth}{!}{
    \begin{tabular}{lccccr}
        \toprule
%         & \multicolumn{2}{c}{Sampling} & & & &                   \\ \cmidrule(lr){2-3}
        Method  & One-shot   &  Modular &  Invertible & Continuous-time &  \\
        \midrule
        JT-VAE &\cmark   &  \cmark     & \xmark & \xmark & \citet{jin2018junction} \\
        MRNN &\xmark   &  \xmark     & \xmark & \xmark & \citet{MRNN} \\
        GraphAF  &\xmark   & \xmark & \cmark & \xmark & \citet{shi2020graphaf}\\
        GraphDF  &\xmark   & \xmark     & \cmark & \xmark & \citet{graphdf} \\
        MoFlow & \cmark  &  \xmark & \cmark & \xmark & \citet{zang2020moflow} \\
        GraphNVP &\cmark   & \xmark     & \cmark & \xmark & \citet{madhawa2019graphnvp} \\
        \midrule
        \atMod & \cmark & \cmark & \cmark & \cmark  & this work \\
        \bottomrule
    \end{tabular}
%    }
\end{table}
% MRNN: generate graph atom-by-atom, using RNN in between
% GraphAF: generate graph atom-by-atom, using AR-flow in between
% GraphDF: generate graph atom-by-atom, uses discrete-flow with real likelihoods and GCN to obtain global graph embeddings. They seem to turn real likelihoods into argmax, but unsure. This is close to us
% MoFlow: GLOW-NF of graph tensors

Earlier models focused on generating sequences based on the SMILES notation \citep{smiles} used in Chemistry to describe the molecular structures as strings. However, they were supplanted by generative models that capture valuable spatial information such as bond strengths and dihedral angles, e.g., by embedding molecular graphs via some graph neural network (GNNs) \citep{GNNsOriginal,GargJJ20}. Such models primarily include variants of Generative Adversarial Networks (GANs), Variational Autoencoders (VAEs), and Normalizing Flows \citep{dinh2014nice,dinh2016density}. Besides known issues with their training, GANs \citep{gan,maziarka2020mol} suffer from the well-documented problem of mode collapse, thereby generating molecules that lack diversity. VAEs \citep{vae,lim2018molecular,jin2018junction}, on the other hand, are susceptible to a distributional shift between the training data and the generated samples. Moreover, optimizing for likelihood via a surrogate lower bound is likely insufficient to capture the complex dependencies inherent in the molecules.

Flows are especially appealing since, in principle, they  enable estimating (and sampling from) complex data distributions using a sequence of invertible transformations on samples from a more tractable continuous distribution. Molecules are discrete, so many flow models \citep{madhawa2019graphnvp,honda2019graph,shi2020graphaf} add noise during encoding and later apply a {\em dequantization} procedure. However, dequantization begets distortion and issues related to convergence \citep{graphdf}. Moroever, many methods segregate the generation of atoms from bonds, so the decoded structure is often not a valid molecule and requires { \em post hoc} correction to ensure validity \citep{zang2020moflow}, effecting a discrepancy between the encoding and the decoded distributions. Permutation dependence is another undesirable artifact of these methods. Some alternatives have been explored to avoid dequantization, e.g.,  \citep{lippe2021categorical} encodes molecules in a continuous latent space via variational inference and jointly optimizes a flow model for generation. Discrete graph flows \citep{graphdf} also circumvent the many pitfalls of dequantization by resorting to discrete latent variables, and performing validity checks during the generative process. However, discrete flows follow an autoregressive procedure that requires a specific ordering of nodes and edges during training. In general, one shot methods can generate much faster than discrete flows. 

We offer a different flow-based perspective tailored to molecules. Specifically, we suggest coupled continuous normalizing E(3)-equivariant flows that bestow generative capabilities from neural partial differential equation (PDE) models on graphs. Graph PDEs have been known to enable designing new embedding methods such as variants of GNNs \citep{chamberlain2021grand}, extending GNNs to continuous layers as Neural ODEs \citep{poli2019graph}, and accommodating spatial information \citep{iakovlev2020learning}. We instead seek to bring to the fore their efficacy and elegance as tools to help generate complex objects, such as molecules, viewed as outcomes resulting from an interplay of co-adapting latent trajectories (i.e., underlying dynamics). Concretely, a flow is associated with each node of the graph, and these flows are conjoined as a joint ODE system conditioned on neighboring nodes. While these flows originate independently as samples from simple distributions, they adjust progressively toward more complex joint distributions as they repeatedly interact with the neighboring flows. We view molecules as samples generated from the globally aligned distributions obtained after many such local feedback iterations. We call the proposed method Modular Flows (\texttt{ModFlow}s) to underscore that each node can be regarded as a module that coordinates with other modules. Table~\ref{tab:relatedworks} summarizes the capabilities of \atMod~compared to some previous generative works.

\paragraph{Contributions.} We propose to learn continuous-time, flow based generative models, grounded on graph PDEs, for generating molecules without resorting to any validity correction. In particular,
\begin{itemize}
    \item we propose \atMod, a novel generative model based on coupled continuous normalizing E(3)-equivariant flows. \atMod~ encapsulates essential inductive bias using PDEs, and defines multiple flows that interact locally toward a globally consistent joint density;  
    \item we encode permutation, translation, rotation, and reflection equivariance with E(3) equivariant GNNs adapted to molecular generation, and can leverage 3D geometric information;
    \item \atMod~ is end-to-end trainable, non-autoregressive, and obviates the need for any external validity checks or correction;
%\item We show that the method can be extended to tree-based and representation and 3D molecules, by utilizing the 3D geometric information. 
    \item empirically, \atMod~ achieves state-of-the-art performance on both the standard QM9~\citep{ramakrishnan2014quantum} and ZINC250K~\citep{irwin2012zinc} benchmarks.%\footnote{Code is available at  }.
\end{itemize}

%Table~\ref{tab:relatedworks} summarizes the capabilities of our model compared to previous works. To be more specific, our contribution is as follows

%\paragraph{Contributions} In this paper, we propose to learn continous-time, flow based generative model based on graph PDEs for generating valid molecular graphs. Our contributions include:

%\begin{itemize}
%    \item We propose \atMod, a novel coupled continuous normalizing E(3)-equivariant flows, based on neural graph PDEs, that repeatedly reconcile locally toward globally aligned densities
%    \item We encode essential inductive bias for molecules using PDEs, which allows us to reconcile locally toward globally aligned densities leading to generate more valid molecules
%    \item We encode permutation, translation, rotation, and reflection equivariance, with E(3) equivariant GNNs adapted to molecular generation
%    \item It is an end-to-end trainable, non-autoregressive, and does not rely on external validity correction techniques.
%    \item We show that the method can be extended to tree-based and representation and 3D molecules, by utilizing the 3D geometric information. 
%    \item Empirically, our model significantly outperforms past methods in various tasks and achieves state of the art on both the standard QM9~\citep{ramakrishnan2014quantum} and ZINC250K~\citep{irwin2012zinc} benchmarks.
%\end{itemize}
\color{black}

%% problem + importance
% generative models are major domain of ML
% . graph generation is an active and still open subdomain
% .. these models can use flows, diffusion, *AEs, GANs or sequential generation 
% ... all of these suffer from \emph{global state}: they evolve huge codes
% we present a local model with tons of benefits

\begin{figure}[t]
    \centering
%    \hspace{-15pt}
    \includegraphics[scale=0.35]{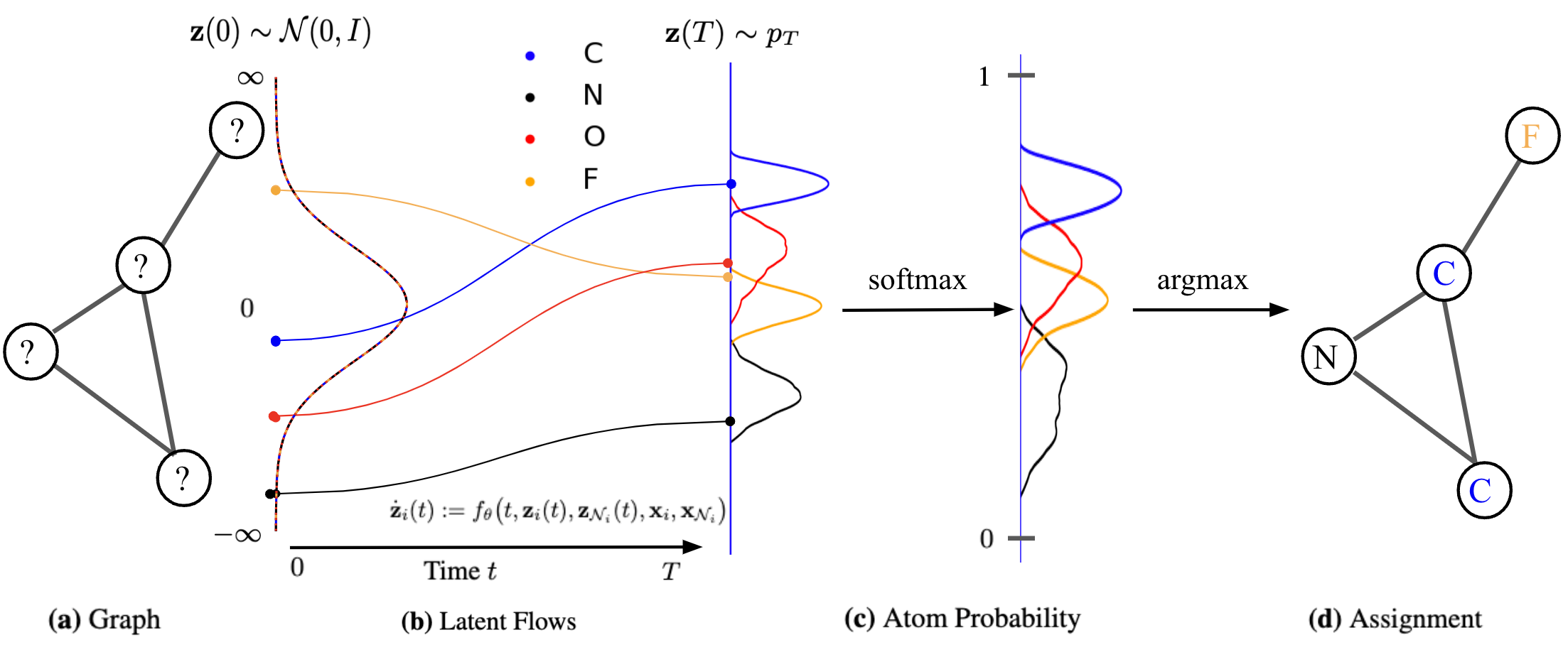}
    \caption{A demonstration of the modular flow generation. The initial Gaussian distributions $\N(0,I)$ evolve into complex densities $\z(T)$ under $f$ and are subsequently translated  into probabilities and labels.}
    \label{fig:workflow}
\end{figure}

%\begin{figure}
%    \centering
%    \includegraphics[scale=0.26]{CNF_NIPS_Paper/combined.png}
%    \caption{Caption}
%    \label{fig:total_workflow}
%\end{figure}

\section{Related works}

\paragraph{Generative models.} 
Earlier attempts for molecule generation \citep{kusner,dai} aimed at representing molecules as SMILES strings \citep{smiles} and developed sequence generation models. A challenge for these approaches is to learn complicated grammar rules that can generate syntactically valid sequences of molecules. Recently, representing molecules as graphs has inspired new deep generative models for molecular generation \citep{segler2018generating,samanta2018designing,neil2018exploring}, ranging from VAEs \citep{jin2018junction,kajino2019molecular} to flows \citep{madhawa2019graphnvp,graphdf,shi2020graphaf}. The core idea is to learn to encode molecular graphs into a latent space, and subsequently decode samples from this latent space to generate new molecules \citep{atwood2016diffusion,xhonneux2020continuous,you2018graphrnn}.

\iffalse 
\paragraph{Normalizing flows.}
In addition to the above methods, normalizing flows have made a significant progress and have been used in various generation tasks (\citet{dinh2016density},\citet{dinh2014nice}). It defines an invertible transformations between latent variables and data samples whi h allows the calculation of the exact data likelihood. Several methods have been proposed which generate molecular graphs by modeling node and adjacency matrices in a one-shot fashion via flow model like GraphNVP \citep{madhawa2019graphnvp}, MoFlow \citep{zang2020moflow} and GRF \citep{honda2019graph}. GraphAF \citep{shi2020graphaf} and GraphDF \citep{graphdf} generates molecular graphs via sequential decision process by adding nodes and edges using auto-regressive and discrete flows.
\fi 

\paragraph{Graph partial differential equations.}
 Graph PDEs is an emerging area that studies PDEs on structured data encoded as graphs. For instance, one can define a PDE on graphs to track the evolution of signals defined over the graph nodes under some dynamics. Graph PDEs have enabled, among others, design of new graph neural networks; see, e.g., works such as  GNODE \citep{poli2019graph}, NeuralPDE \citep{iakovlev2020learning}, Neural operator \citep{li2020neural}, GRAND \citep{chamberlain2021grand}, and PDE-GCN \citep{eliasof2021pde}. Different from all these works, we focus on using PDEs for generative modeling of molecules (as graph-structured objects). Interestingly, ~\atMod ~ proposed in this work may be viewed as a new equivariant temporal graph network \citep{temporal, PINT2022}.

% Different from these works, we focus on generative modeling by viewing molecules as samples from a (possibly complex) distribution, arising as a result of coupling node flows. 

% this distribution  coupled flows cast as a PDE system.   
 
%  Generating molecular structures may be studied as a dynamical system \citep{pde_mol}:  samples from an initial distribution evolve under some dynamics toward some a distribution,  which we output the generated molecule and its atoms and edges.  %Moreover, one can combine these coupled PDE approaches with diffusion models to capture essential correlations among variables leading to supreme performance of the generative model \citep{pde_mol}.

\paragraph{Validity oracles.} 
A key challenge of molecular generative models is to be able to generate valid molecules, according to various criteria for molecular validity or feasibility. It is a common practice to call on external chemical software as rejection oracles to reduce or exclude invalid molecules, or do validity checks as part of autoregressive generation \citep{graphdf,shi2020graphaf,MRNN}. An important open question has been whether generative models can learn to achieve high generative validity {\em intrinsically}, i.e., without being aided by oracles or resorting to additional checks. ~\atMod~ takes a major step forward toward that goal.

\section{Modular Flows}

We focus on unsupervised learning of an underlying graph density $p(G)$ using a dataset of observed molecular graphs $\D = \{G_n\}_{n=1}^N$. We learn a generative flow model $p_\theta(G)$ specified by flow parameters $\theta$, and use it to sample novel high-probability molecules.

\subsection{Molecular Representation}

\paragraph{Graph representation.} We represent each molecular graph $G = (V,E)$ of size $M$ as a tuple of vertices $V = (v_1, \ldots, v_M)$ and edges $E \subset V \times V$. Each vertex takes a value from an alphabet on atoms:  $v \in \A = \{ \atC,\atH,\atN,\atO,\atP,\atS,\ldots \}$; while each edge $e \in \B = \{1,2,3\}$ abstracts some bond type (i.e., single, double, or triple). We assume that, conditioned on the edges, the graph likelihood factorizes as a product of categorical distributions over vertices given their latent representations:
\begin{align} \label{eq:3}
    p(G) &:= p(V | E,\{ z\}) = \prod_{i=1}^M \Cat(v_i | \sigma(\z_i))~, 
\end{align}
where $\z_i = (z_{i\atC}, z_{i\atH}, \ldots) \in \R^{|\A|}$ is a set of atom scores for node $i$ such that $z_{ik} \in \R$ pertains to type $k \in \A$, and $\sigma$ is the softmax function
\begin{align}
    \sigma(\z_i)_k &= \frac{\exp (\z_{ik}) }{\sum_{k'} \exp (\z_{ik'})}~,
\end{align}
which turns the real-valued scores $\z_i$ into normalized probabilities. ~\atMod~ also  supports 3D molecular graphs that contain atomic coordinates and angles as additional information.

\paragraph{Tree representations.}
We can obtain an alternative representation for molecules: we can decompose each molecule into a tree-like structure, by contracting certain vertices into a single node (denoted as a cluster) such that the molecular graph becomes acyclic. Following \citet{jin2018junction}, we restrict these clusters to ring substructures present in the molecular data, in addition to the atom alphabet. Thus, we obtain an extended alphabet $\A_{\mathrm{tree}} = \A \cup \{\texttt{C}_{1},\texttt{C}_{2},\ldots \}$, where each cluster label $\texttt{C}_{r}$ corresponds to some ring substructure in the label vocabulary $\rchi$. We then reduce the vocabulary to the $30$ most commonly occurring substructures of $\A_{\mathrm{tree}}$. For further details, see Appendix~\ref{sec:tree}. 

% [Markus: I removed this detail, maybe this could go to supplement]
%The vocabulary has a limited size ($|\rchi| \approx 780$), but it can be empirically observed that it follows a skewed distribution over frequency of appearance within dataset. It means that only a subset ($\sim 30$) of ring substructures (labels) have high frequency of appearance as compared to other substructures within the $\rchi$. Considering the above observation, we only consider $\sim 30$ high frequency cluster labels from vocabulary to be a part of the set $\A$. This represents as an expansion to our current representation described in Eq.~\ref{eq:3} and forms an extension of modular flows on tree representation. 

\subsection{Differential modular flows}

Normalizing flows \citep{NF} provide a general recipe for constructing flexible probability distributions, used in density estimation \citep{density1,density2} and generative modeling \citep{gen1,zang2020moflow}. We propose to model the atom scores $\z_i(t)$ as a Continuous-time Normalizing Flow (CNF) \citep{grathwohl2018ffjord} over time $t \in \R_+$. We assume the initial scores at time $t=0$ follow an uninformative Gaussian base distribution $\z_i(0) \sim \N(0,I)$ for each node $i$. Node scores evolve in parallel over time according to the differential equation
\begin{align}
    \dz_i(t) &:= \frac{\partial \z_i(t)}{\partial t} = f_\theta\big( t, \z_i(t), \z_{\N_i}(t),\mathbf{x}_{i}, \mathbf{x}_{\N_i} \big), \qquad i \in \{1, \ldots, M\}~,
\end{align}
where $\N_i = \{j: (i,j) \in E \}$ is the set of neighbors of node $i$ and $\z_{\N_i}(t) = \{\z_j(t): j \in \N_i \}$ the scores of the neighbors at time $t$; $\mathbf{x}_{i}$ and  $\mathbf{x}_{\N_i}$ denote, respectively, the positional (2D/3D) information of $i$ and its neighbours; and $\theta$ denotes the parameters of the flow function $f$ to be learned. Stacking together all node differentials, we obtain a \emph{modular} system of coupled ODEs:
\begin{align} \label{eq:dz}
    \dz(t) &= \begin{pmatrix} \dz_1(t) \\ \vdots \\ \dz_M(t) \end{pmatrix} = \begin{pmatrix} f_\theta\big( t, \z_1(t), \z_{\N_1}(t),\mathbf{x}_{i}, \mathbf{x}_{\N_i} \big) \\ \vdots \\ f_\theta\big( t, \z_M(t), \z_{\N_M}(t),\mathbf{x}_{i}, \mathbf{x}_{\N_i} \big) \end{pmatrix} \\
    \z(T) &= \z(0) + \int_0^T \dz(t) dt~.
\end{align}
This coupled system of ODEs may be viewed as a graph PDE \citep{iakovlev2020learning,chamberlain2021grand}, where the evolution of each node depends only on its neighbors. 

The joint flow induces a corresponding change in the individual densities in terms of divergence of $f$ \citep{chen2018neural},
\begin{align} \label{eq:flow_prob}
    \frac{d \log p_t(\z_i(t))}{d t} &= - \tr \left( \frac{\partial f_\theta\big(t, \z_i(t), \z_{\N_i}(t),\mathbf{x}_{i}, \mathbf{x}_{\N_i}\big)}{\partial \z_i} \right),
\end{align}
starting from the base distribution $p_0(\z_i(0)) = \N(\z_i(0) | 0,I)$. The trace picks only the diagonal elements of the Jacobian $\frac{\partial f}{\partial \z}$, which interprets the input from neighbors, $\z_{\N_i}$, as a `control' for each node $\z_i$ at each instant $t$. An ODE solver is used for such systems, and the gradients are computed via the adjoint sensitivity method \citep{kolmogorov1962probability}. This approach incurs a low memory cost, and explicitly controls the numerical error. Notably, moving towards modular flows translates sparsity also to the adjoints.

\paragraph{Proposition 1:}\emph{Modular adjoints are sparser than regular adjoints.  They can be computed as } \\
%We note that the modular continuous-time flow gradients can be computed by the adjoint method, which results in modular adjoints $\bl_i(t) := \frac{\partial \mathcal{L}}{\partial \z_i}$. That is, we have
\begin{align}
    \frac{d\bl_i}{dt} &= - \bl(t)^{\top} \frac{\partial f\big(t, \z_i(t), \z_{\N_i}(t),\mathbf{x}_{i}, \mathbf{x}_{\N_i}\big) }{\partial \z} = - \sum_{j \in \N_i \cup \{i\}} \bl_j(t)^{\top} \frac{\partial f\big(t, \z_i(t), \z_{\N_i}(t),\mathbf{x}_{i}, \mathbf{x}_{\N_i}\big) }{\partial \z_j},
\end{align}
\emph{where the partial derivatives $\frac{\partial f}{\partial \z} = [\frac{\partial f_i}{\partial \z_j}]_{ij}$ are sparse} (see Appendix~\ref{sec:adjoint} for the derivation).

\subsection{Equivariant local differential}

Our goal is to have a differential function $f$ that is a PDE operator used in ~\autoref{eq:dz}, and that satisfies the natural equivariances and invariances of the molecules. Specifically, this function must be (i) translation equivariant: translating the input results in an equivalent translation of the output; (ii) rotational (and reflection) equivariant: rotating the input results in an equivalent rotation of the output; and (iii) permutation equivariant: permuting the input results in the same permutation of the output. Therefore, we chose to use E(3)-Equivariant GNN (EGNN) ~\citep{egnn}, which is translation, rotation and reflection equivariant (E(n)), and permutation equivariant with respect to an input set of points (see Appendix~\ref{sec:egnn} for details). EGNN takes as input  the node embeddings as well as the geometric information (polar coordinates (2D) and spherical polar coordinates (3D)). Interestingly, \atMod~ can be viewed as a message passing temporal graph network \citep{temporal, PINT2022} as shown next. 
\paragraph{Proposition 2:}\emph{Modular Flows can be cast as message passing Temporal Graph Networks (TGNs). The operations are listed in Table~\ref{tab:egnn_tgn}, where {\rm \atMod~} is subjected to a single layer of EGNN.}  (See Appendix~\ref{sec:temporal_to_egnn} for more details). 
%We note that the modular continuous-time flow gradients can be computed by the adjoint method, which results in modular adjoints $\bl_i(t) := \frac{\partial \mathcal{L}}{\partial \z_i}$. That is, we have
\begin{table}[!h]
\centering
    \caption{\atMod~ as a temporal graph network (TGN). Adopting notation for TGNs from \cite{temporal} $v_i$ is a node-wise event on $i$; $e_{ij}$ denotes an (asymmetric) interaction between $i$ and $j$; $\mathbf{s}_{i}$ is the memory of node $i$; and $t$ and $t^-$ denote time with $t^-$ being the time of last interaction before $t$,  e.g., $\mathbf{s}_{i}(t^-)$ is the memory of $i$ just before time $t$; and $\operatorname{msg}$ and $\operatorname{agg}$ are learnable functions (e.g., MLP) to compute, respectively, the individual and the aggregate messages. For \atMod, we use   $\mathbf{r}_{ij}$ to denote the spatial distance  $\mathbf{x}_{i} - \mathbf{x}_{j}$, and $a_{i j}$ to denote the attributes of the edge between $i$ and $j$. The functions $\phi_e$, $\phi_x$, and $\phi_h$ are as defined in \citep{egnn}.  \label{tab:egnn_tgn}}
    \centering
    \resizebox{\textwidth}{!}{
    \begin{tabular}{lcc}
        \toprule
        Method  & TGN layer &  \atMod   \\
        \midrule
        
        \multirow{2}{4em}{Edge}  & $\mathbf{m}_{i j}'(t)=\operatorname{msg}\left(\mathbf{s}_{i}\left(t^{-}\right), \mathbf{s}_{j}\left(t^{-}\right),  \Delta t, \mathbf{e}_{ij}(t)\right)  $ & $\mathbf{m}_{i j}(t) =\phi_{e}\left(\mathbf{z}_{i}(t), \mathbf{z}_{j}(t),\left\|\mathbf{r}_{ij}(t)\right\|^{2}, a_{i j}\right)$   \\
        
         & $\overline{\mathbf{m}}_{i}'(t)=\operatorname{agg}\left(\{\mathbf{m}_{i j}'\left(t\right) | j \in  \N_i \}\right) $ & $\mathbf{m}_{i}(t) = \sum_{j\in \mathcal{N}(i)} \mathbf{m}_{ij} $   \\
         & & $\hat{\mathbf{m}}_{ij}(t) = \mathbf{r}_{ij}(t)\cdot \phi_{x}\left(\mathbf{m}_{i j}(t)\right) $ \\
         &  & $\hat{\mathbf{m}}_{i}(t) = C \sum_{j\in \mathcal{N}(i)} \hat{\mathbf{m}}_{ij}(t) $   \\
         \midrule
        Memory state  & $\mathbf{s}_{i}(t)=\operatorname{mem}\left(\overline{\mathbf{m}}_{i}'(t), \mathbf{s}_{i}\left(t^{-}\right)\right)$  &$\mathbf{x}_{i}(t+1) = \mathbf{x}_{i}(t)+ \hat{\mathbf{m}}_{i}(t)$   \\
        \midrule
        Node  & $\mathbf{z}_{i}'(t)=\sum_{j \in \N_i} h\left(\mathbf{s}_{i}(t), \mathbf{s}_{j}(t), \mathbf{e}_{i j}(t), \mathbf{v}_{i}(t), \mathbf{v}_{j}(t)\right)$ & $\mathbf{z}_{i}(t+1) =\phi_{h}\left(\mathbf{z}_{i}(t), \mathbf{m}_{i}(t)\right)$  \\
        \bottomrule
    \end{tabular}}
\end{table}

%\subsection{Invariant local polar differential}

%Our goal is to have a differential function $f$ that is a PDE operator, and which %satisfies invariances:
%\begin{itemize}
%    \item Translation invariance
%    \item Rotational invariance
%    \item Permutation invariance
%    \item Size invariance. [Add definitions?]
%\end{itemize}
%We first propose to represent the local neighborhood of central node $v_i$ as a %polar distribution of angle $\a \in \R^{D-1}$ and radius $r \in \R$ over time $t$,
%\begin{align}
%    p_i(\a,r)(t) &\sim \frac{1}{N} \sum_j \underbrace{\Delta \z_j(t)}_{\z_i - \z_j %\in \R^{|\A|}} \N\left( \begin{pmatrix} \a_j \\ r_j \end{pmatrix}, \begin{pmatrix} %\boldsymbol{\sigma}_a & 0 \\ 0 & \sigma_r \end{pmatrix} \right),
%\end{align}
%and convolve it using a CNN
%\begin{align}
%    f_\theta\big( t, \z_i(t), \z_{\N_i}(t) \big) %&:= \E[ g_{t,\theta} \star %p_i(t) ] \\
%    &:= \mathrm{TCN}(p_i(t) ; \theta, t)
%\end{align}
%where $\sigma_r,\boldsymbol{\sigma}_a$ are the variance hyperparameters, and %$\mathrm{TCN}$ is a temporal convolution network [CITE] (See appendix X).
\subsection{Training objective}
Normalizing flows are predominantly trained to minimize $\kl[p_{\mathrm{data}} || p_\theta]$, i.e., the KL-divergence  between the unknown data distribution $p_{\mathrm{data}}$ and the flow-generated distribution $p_\theta$. This objective is equivalent to maximizing $\E_{p_{\mathrm{data}}}[\log p_\theta]$ \citep{papamakarios2021normalizing}. However, note that the discrete graphs $G$ and the continuous atom scores $\z(t)$ reside in different spaces. Thus, in order to apply flows, a mapping between the observation space and the flow space is needed. Earlier approaches use dequantisation to turn a graph $G$ into a distribution of latent states, and argmax to deterministically map latent states to graphs \citep{zang2020moflow}.
\begin{figure}[h]
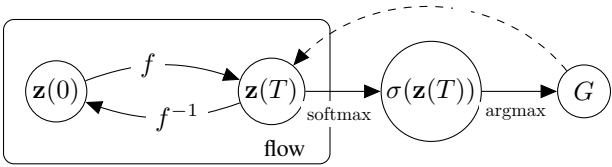

\begin{center}
\tikz{
    \node[latent] (z0) {$\z(0)$}; 
%    \node[latent,xshift=5mm,right=of z0] (zt) {$\z(t)$};
    \node[latent,xshift=10mm,right=of z0] (zT) {$\z(T)$}; 
    \node[latent,right=of zT] (sT) {$\sigma(\z(T))$}; 
    \node[latent,right=of sT] (G) {$G$}; 
    \plate [inner sep=.3cm,xshift=.02cm,yshift=.2cm] {flow} {(z0) (zT)} {flow};
    \path[->] (z0) edge[bend left=20] node[fill=white, anchor=center, pos=0.4, scale=1.0] {$f$} (zT);
%     \path[->] (zt) edge[bend left=20] node[fill=white, anchor=center, pos=0.4, scale=1.0] {$f$} (zT);
%    \path[->] (zT) edge[bend left=20] node[fill=white, anchor=center, pos=0.4, scale=1.0] {$f^{-1}$} (zt);
    \path[->] (zT) edge[bend left=20] node[fill=white, anchor=center, pos=0.4, scale=1.0] {$f^{-1}$} (z0);
    \path[->] (zT) edge node[fill=white, anchor=center, yshift=-3mm, pos=0.45, scale=0.7] {$\mathrm{softmax}$} (sT);
    \path[->] (sT) edge node[fill=white, anchor=center, yshift=-3mm, pos=0.45, scale=0.7] {$\mathrm{argmax}$} (G);
    \path[->,dashed] (G) edge[bend right=50] node[fill=white, anchor=center, pos=0.4, scale=1.0]{}(zT);
    
}
\end{center}
\caption{Plate diagram showing both the inference and generative components of \atMod.}
\label{fig:plate}
\end{figure}

%A common objective of normalizing flow density estimators is to minimize the KL divergence
%\begin{align}
%    \argmin_\theta \quad \kl[ p_{\mathrm{data}}(G) \, || \, p_\theta(G) ]
%\end{align}
%between the data distribution $p_\mathrm{data}$ and the flow-induced graph density $p_\theta(G) = \int p(G|\z(T)) p_T(\z(T) | \theta) d\z(T)$ marginalises likelihood contributions towards $G$ from the entire score distribution $p_T$ (See Figure XX). In contrast, the argmax approach is based on Dirac matching $\delta(G|\mathrm{argmax} \, \z(T))$. This is equivalent to maximizing the \emph{marginal likelihood}

We instead reduce the learning problem to maximizing $\E_{\hat{p}_{\mathrm{data}}(\z(T))}[\log p_\theta(\z(T))]$, where we turn the observed set of graphs $\{G_n\}$ into a set of scores $\{\z_n\}$ using
%$\z_n := (1-\epsilon) \mathrm{onehot}(G_{n}) + \epsilon$ ($ \texttt{onehot}_{\epsilon}$)
$$\z_n (G_n; \epsilon) = (1-\epsilon)~\mathrm{onehot}(G_n) ~+~ \dfrac{\epsilon}{|\mathcal{A_s}|} \textbf{1}_{M(n)} \textbf{1}_{|\mathcal{A_s}|}^{\top}~,$$
where $\mathrm{onehot}(G_n)$ is a matrix of size $M(n) \times |\mathcal{A_s}|$ (i.e., rows equal to the number of nodes in $G_n$ and columns equal to the number of possible node labels) such that $G_n(i, k)$ = 1 if $v_i = a_k \in \mathcal{A_s}$, that is if the vertex $i$ is labeled with atom $k$, and 0 otherwise;  
$\textbf{1}_{q}$ is a vector with $q$ entries each set to 1; $\mathcal{A_s} \in \{\mathcal{A}, \mathcal{A}_{\rm tree} \}$; and $\epsilon \in [0,1]$ is added to model the noise in estimating the posterior $p({\z(T)|G})$ due to short-circuiting the inference process from $G$ to $\z(T)$ skipping the intermediate dependencies, thereby inducing an unconditional distribution $\hat{p}_{\mathrm{data}}$ that is slightly different from the true data distribution $p_{\mathrm{data}}$. The plate diagram in \autoref{fig:plate} summarizes the overall procedure. 

Effectively, we exploit the (non-reversible) composition of the argmax and softmax operations to transition from the continuous flow space to the discrete graph space, but skip this composition altogether in the reverse direction.  Importantly, this short-circuiting allows \atMod~ to keep the forward and backward flows between $\z(0)$ and $\z(T)$ completely aligned (i.e., reversible) unlike previous approaches. We maximize the following objective over $N$ training graphs:
\begin{align}\label{eq:objective}
    \arg\!\max_\theta \L &= \E_{\hat{p}_{\mathrm{data}}(\z)} \log p_\theta(\z) \\
    &\approx \frac{1}{N} \sum_{n=1}^N \log p_T\big( \z(T) = \z_n \big) \\
    &= \frac{1}{N} \sum_{n=1}^N \left( \sum_{i=1}^{M(n)} \log p_0(\z_i(0)) - \sum_{i=1}^{M(n)} \int_0^T \mathrm{tr} \frac{\partial f_\theta(t, \z_i(t), \z_{\N_i}(t),\mathbf{x}_{i}, \mathbf{x}_{\N_i})}{\partial \z_i(t)} dt\right)~,
\end{align}
which factorizes over the size $M(n)$ of the $n$'th training molecule. The encoding probability follows from \autoref{eq:flow_prob}, where $\z(0)$ can be traced by traversing the flow $f$ backward in time starting from $\z_n$ at time $t=T$ until $t=0$. In practice we solve ODE integrals using a numerical solver such as Runge-Kutta. We thus delegate this task to a general solver of the form $\texttt{ODESolve}(\z,f_{\theta},T)$, where 
map $f_{\theta}$ is applied for $T$ steps starting with $\z$. An optimizer $\texttt{optim}$ is also required for updating $\theta$. 

\subsection{Molecular generation}
Given a molecular structure, we can generate novel molecules by sampling an initial state $\z(0) \sim \N(0,I)$, and running the modular flow forward in time for $T$ steps and obtain $\z(T)$. This procedure maps a tractable base distribution $p_0$ to a more complex distribution $p_T$. %We implemented different variants on 2D, 3D and tree representation corresponding to molecular graph generation tasks. 
We follow argmax to pick the most probable label assignment for each node \citep{zang2020moflow}. We outline the procedures for training and generation in Algorithm~\ref{algorithm} and Algorithm~\ref{gen_algorithm} respectively.
%Our results indicate that the learned $p_T$ is diverse and almost entirely valid. [\textcolor{red}{Should we put in Appendix?Might increase length beyond 9 pages}]
\algrenewcommand\algorithmicindent{1.0em}
\begin{figure}[bt]
\begin{multicols}{2}
\begin{algorithm}[H]
    \caption{Training \atMod} \label{algorithm}
    \begin{algorithmic}[1]
    \Require Dataset $\mathcal{D}$, iterations $n_{\mathrm{iter}}$, batch size $B$, number of batches $n_B$ 
    \State Initialise parameters $\theta$ of \atMod~ (EGNN)
    \For{$k = 1, \ldots, n_{\mathrm{iter}}$}
        \For{$b = 1,\ldots,n_B$} 
            \State Sample $\D_b = \{G_1,\ldots,G_B\}$ from $\D$
            \State Define $\z_b(T) := \{\z_r(T) : G_r \in \D_b\}$
            \State Set $\z_b(T)$ to $\z_b (G_b; \epsilon)$ %$\texttt{one_hot}(\mathbf{G})$
            \State $\L_b =  \dfrac{1}{B} \displaystyle \sum_{G_r \in D_b} \log p_{\theta} (\z_r(T))$%\log p_\theta(G_b) $
            , using\qquad \hspace*{0.7cm}
             $\z_b(0) = \texttt{ODESolve}(\z_b(T),f_{\theta}^{-1},T)$
            \EndFor
            \State $\theta \xleftarrow{} \texttt{optim}(\frac{1}{n_B} \sum_{b=1}^{n_B} \L_b; \theta)$
    \EndFor
    \end{algorithmic}
\end{algorithm}
\columnbreak
\begin{algorithm}[H]
    \caption{Generating with \atMod}\label{gen_algorithm}
    \begin{algorithmic}[1]
%    \Require Connectivity $E$
        \State Sample $\z(0) \sim \mathcal{N}(0,I)$
        \State $\z(T) = \texttt{ODESolve}(\z(0),f_{\theta},T)$
        \State Assign labels by  $\mathrm{argmax}(\sigma(\z(T)))$
    \end{algorithmic}
\end{algorithm}
\end{multicols}
\end{figure}
\section{Experiments}
We first demonstrate the ability of Modular Flows (\atMod) to learn highly discontinuous synthetic patterns on 2D grids.  We also evaluated \atMod~ models trained, variously, on (i) 2D coordinates, (ii) 3D coordinates, (iii) 2D coordinates + tree representation, and (iv) 3D coordinates + tree representation on the tasks of molecular generation and optimization. Our results show that \atMod~  compares favorably to other prominent flow and non-flow based molecular generative models, including GraphDF \citep{graphdf}, GraphNVP \citep{madhawa2019graphnvp}, MRNN\citep{MRNN}, and GraphAF \citep{shi2020graphaf}. Notably, \atMod~ achieves state-of-the-art results without validity checks or post hoc correction. We also provide results of our ablation studies to underscore the relevance of geometric features and equivariance toward this superlative empirical performance. 
\iffalse
\begin{figure*}
    %\vspace{-10pt}
    %\hspace{-20pt}
    \centering
    \includegraphics[width=0.5\textwidth,height=4.5cm,trim={0 0.1cm 0 0.3cm}, clip]{toy_final.png}
    \caption{\atMod~ can accurately learn to reproduce complex, discontinuous graph patterns.}
    \label{fig:toy_data}
\end{figure*}
\fi
\subsection{Density Estimation}
%\begin{wrapfigure}{r}[13]{0.6\textwidth}
%    %\vspace{-10pt}
%    \hspace{-10pt}
%    \centering
%    \includegraphics[width=0.5\textwidth,trim={0 0.1cm 0 0}, clip]{toy_final.png}
%    \caption{\atMod~ can accurately learn to reproduce complex, discontinuous graph patterns.}
%    \label{fig:toy_data}
%\end{wrapfigure}
\begin{wrapfigure}[15]{r}{0.6\textwidth}
    \vspace{-10pt}
    \hspace{-9pt}
    \includegraphics[scale=0.24, trim={0 0.1cm 0 0}, clip]{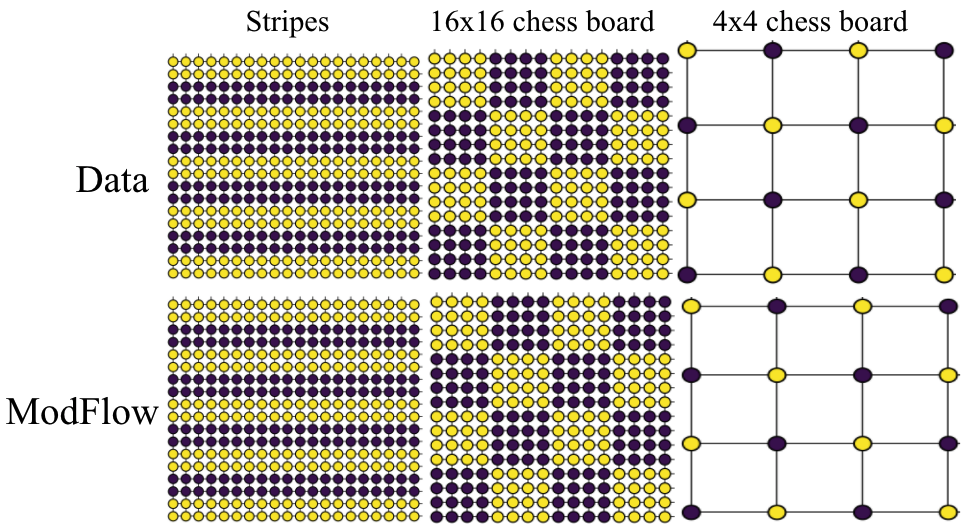}
    \caption{\atMod~ can accurately learn to reproduce complex, discontinuous graph patterns.}
    \label{fig:toy_data}
\end{wrapfigure}
We generated our synthetic data in the following way. We considered two variants of a chessboard pattern, namely, (i) $4 \times 4$ grid where every node takes a binary value, 0 or 1, and neighboring nodes have different values; and (ii) $16 \times 16$ grid where nodes in each block of $4 \times 4$ all take the same value (0 or 1), different from the adjacent blocks. We also experimented with a $20 \times 20$ grid describing alternating stripes of $0$s and $1$s. 

\autoref{fig:toy_data} shows that \atMod~ can learn neural differential functions $f_\theta$ that reproduce the patterns almost perfectly, indicating sufficient capacity to model complex patterns. That is, \atMod~ is able to transform the initial Gaussian distribution into different multi-modal and discontinuous distributions. %We used argmax to deterministically map the latent state to node label. We can also see that our model is capable of modeling disconnected modes and able to learn approximations of discontinuous density functions.

\subsection{Molecule Generation}

\paragraph{Data.} We trained and evaluated all the models on ZINC250k \citep{irwin2012zinc} and QM9 \citep{ramakrishnan2014quantum} datasets. The ZINC250k set contains 250,000 drug-like molecules, each consisting of up to 38 atoms. The QM9 set contains 134,000 stable small organic molecules with atoms from the set $\{\atC, \atH, \atO, \atN, \atF\}$. The molecules are processed to be in the {\em kekulized} form with hydrogens removed by the RDkit software \citep{landrum2013rdkit}.
\begin{table}[!htp]
    \caption{Random generation on QM9 (top) and ZINC250K (bottom) without post hoc validity corrections. Results with $^*$ are taken from \citet{graphdf}. Higher values are better for all columns.}
    \label{tab:results1}
    \centering
    \resizebox{0.88\textwidth}{!}{
    \begin{tabular}{lcccc}
        \toprule
        Method & Validity \%  & Uniqueness \%& Novelty \% &  Reconstruction \%  \\
        \midrule
        GVAE        & 60.2 & 9.3 & 80.9 & 96.0 \\
        GraphNVP$^*$ & 83.1 & 99.2  &  58.2    & 100 \\
        GRF$^*$ &  84.5 & 66  &  58.6     & 100  \\
        GraphAF$^*$  &67 & 94.2  &  88.8     & 100  \\
        GraphDF$^*$  & 82.7 & 97.6  &  98.1     & 100  \\
        MoFlow$^*$ &   89.0 & 98.5  &  96.4     & 100  \\
        \midrule
        \atMod~ (2D-EGNN)  &$\textbf{96.2} \pm 1.7$ & $\textbf{99.5}$ & $\textbf{100}$ & 100  \\
        \atMod~ (3D-EGNN)  & $\textbf{98.3} \pm 0.7$ & 99.1 & $\textbf{100}$ & 100  \\
        \atMod~ (JT-2D-EGNN) &$\textbf{97.9} \pm 1.2$ &99.2 & $\textbf{100}$ & 100  \\
        \atMod~ (JT-3D-EGNN)  &$\textbf{99.1} \pm 0.8$ &99.3 & $\textbf{100}$ & 100   \\
        \bottomrule
    \end{tabular}
    }
\end{table}
\begin{table}[!htp]
    %\caption{Random generation performance on ZINC250K dataset without post hoc validity corrections. Results with $^*$ are taken from \citet{graphdf}. Higher values are better for all columns.}
    \label{tab:results2}
    \centering
    \resizebox{0.88\textwidth}{!}{
    \begin{tabular}{lcccc}
        \toprule
        Method  &  Validity \%  & Uniqueness \% & Novelty \% &  Reconstruction \% \\
        \midrule
        MRNN        &  65   & 99.89  &  100    & n/a\\
        GVAE        & 7.2 & 9 & 100 & 53.7 \\
        GCPN        & 20   & 99.97    & 100   & n/a       \\
        GraphNVP$^*$ & 42.6 & 94.8  &  100    & 100 \\
        GRF$^*$ & 73.4 & 53.7  &  100     & 100  \\
        GraphAF$^*$  &68 & 99.1  &  100     & 100  \\
        GraphDF$^*$  & 89  & 99.2  & 100     & 100  \\
        MoFlow$^*$ &  50.3 & \textbf{99.9}  &  100     & 100  \\
        \midrule
        \atMod~ (2D-EGNN) & $\textbf{94.8} \pm 1.0$ & 99.4 & 100 & 100  \\
        \atMod~ (3D-EGNN) &  $\textbf{95.4} \pm 1.2$ & 99.7 & 100 & 100 \\
        \atMod~ (JT-2D-EGNN) & $\textbf{97.4} \pm 1.4$ & 99.1 & 100 & 100  \\
        \atMod~ (JT-3D-EGNN) & $\textbf{98.1} \pm 0.9$ & 99.3 & 100 & 100  \\
        \bottomrule
    \end{tabular} }
\end{table}
\paragraph{Setup.} We adopt common quality metrics to evaluate molecular generation. {\em Validity} is the fraction of molecules that satisfy the respective chemical valency of each atom. {\em Uniqueness} refers to the fraction of generated molecules that is unique (i.e, not a duplicate of some other generated molecule). {\em Novelty} is the fraction of generated molecules that is not present in the training data. {\em Reconstruction} is the fraction of molecules that can be reconstructed from their encoding. Here, we strictly limit ourselves to comparing all methods on their validity scores \emph{without resorting to external correction}. We trained each model with 5 random weight initializations, and generated 50,000 molecular graphs for evaluation. We report the mean and the standard deviation scores across these multiple runs.

\textbf{Implementation.} The models were implemented in PyTorch \citep{paszke2019pytorch}. The EGNN method used only a single layer with embedding dimension of 32. %The input concatenates time and scalar vocabulary scores, per node. 
We trained with the Adam optimizer \citep{kingma2014adam} for 50-100 epochs (until the training loss became stable), with batch size 1000 and learning rate $0.001$. \atMod~ is significantly faster compared to autoregressive models such as GraphAF and GraphDF. For more details, see Appendix~\ref{sec:implement}.

\textbf{Results.} \autoref{tab:results1} reports the performance on QM9 (top) and ZINC250K (bottom) respectively. \atMod~ achieves state-of-the-art results across all metrics. Notably, its reconstruction rate is 100\% (similar to other flow models); in addition, however, the novelty (100\%) and uniqueness scores ($\approx$99\%) are also very high. Moreover, \atMod~ surpassed the other methods on validity (95\%-99\%). 

In Appendix~\ref{sec:moses}, we document additional evaluations with respect to the MOSES metrics that access the overall quality of generated molecules, as well as the distributions of chemical properties. All these results substantiate the promise of \atMod~ as an effective tool for molecular generation. 

\begin{figure}[!htp]
\centering
\begin{subfigure}{.5\textwidth}
  \centering
  \includegraphics[width=.95\linewidth]{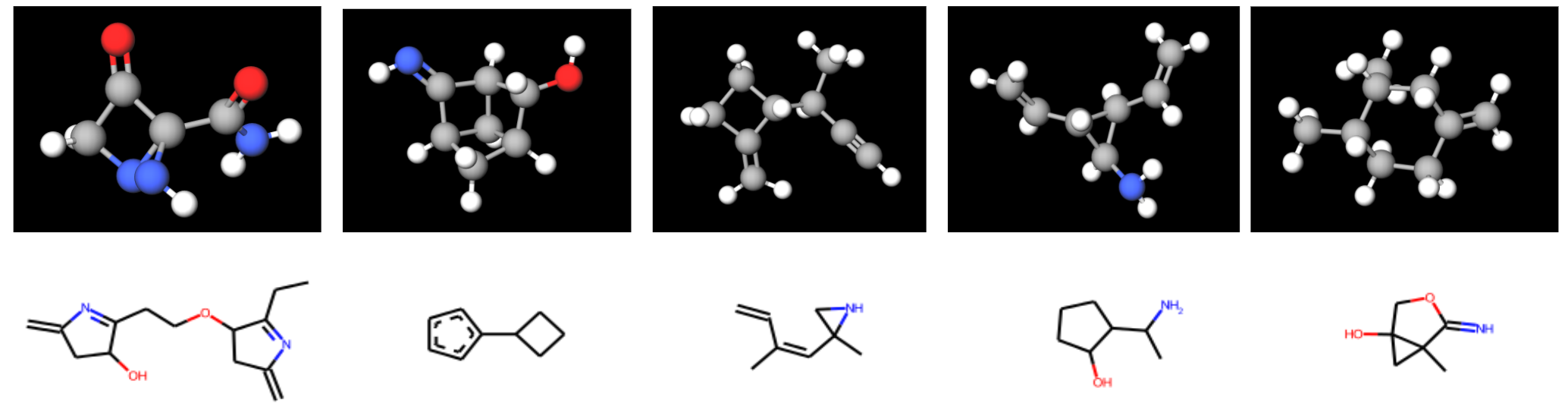}
  \caption{QM9 Dataset}
  \label{fig:gen_mol_qm9}
\end{subfigure}%
\begin{subfigure}{.5\textwidth}
  \centering
  \includegraphics[width=.95\linewidth]{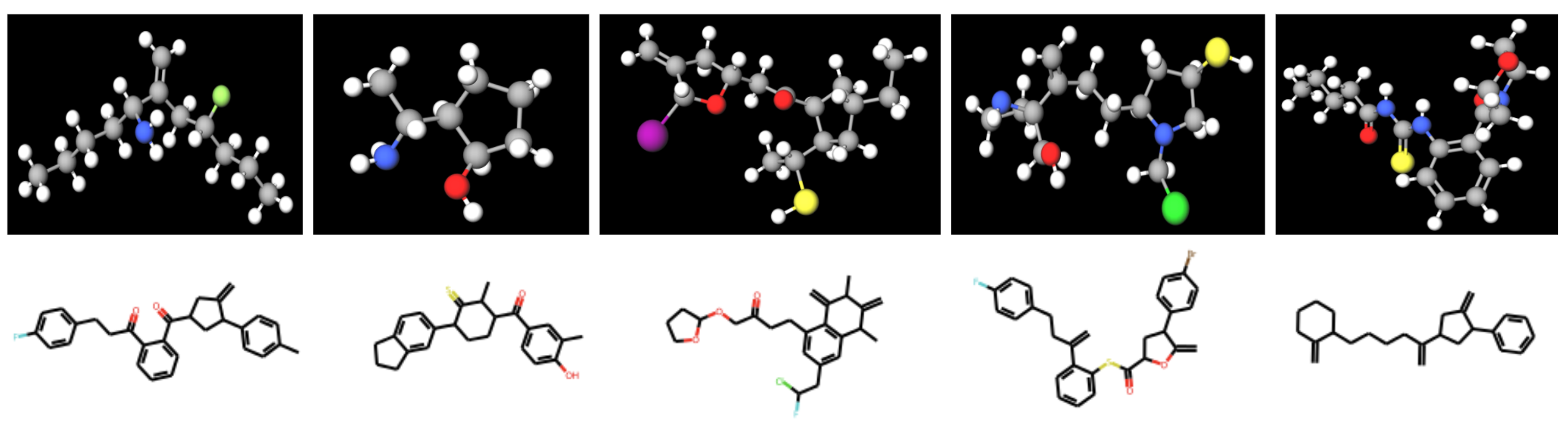}
  \caption{ZINC250K Dataset}
  \label{fig:gen_mol_zinc}
\end{subfigure}
\caption{Samples of  molecules generated by \atMod. More examples are shown in Appendix \ref{sec:GenMols}.}
\label{fig:all_gen_mol}
\end{figure}

\subsection{Property-targeted Molecular Optimization}
\label{sec:prop_opt}
The task of molecular optimization is to search for molecules that have better chemical properties. We choose the standard quantitative estimate of drug-likeness (QED) as our target chemical property. QED measures the potential of a molecule to be characterized as a drug. We used a pre-trained ModFlow model $f$ to encode molecules $\mathcal{M}$ into their embeddings $\mathcal{Z} = f(\mathcal{M})$, and applied linear regression to obtain QED scores $\mathcal{Y}$ from these embeddings.  We then interpolate in the latent space of each molecule along the direction of increasing QED via several gradient ascent steps, i.e., updates of the form $\mathcal{Z}' = \mathcal{Z} + \lambda*\frac{d\mathcal{Y}}{d\mathcal{Z}}$, where  $\lambda$ denotes the length of the search step. The final embedding thus obtained is decoded as a new molecule via the reverse mapping $f^{-1}$. 

\begin{figure}[!htp]
    \centering
    \includegraphics[scale=0.4]{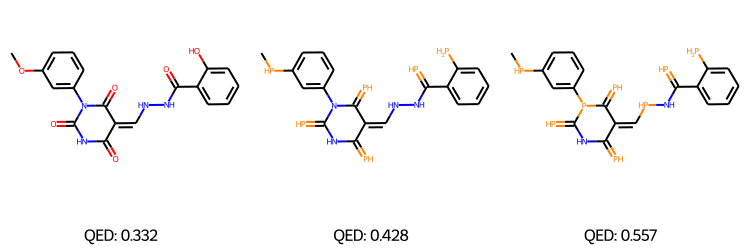}
    \caption{Example of chemical property optimization on the ZINC250K dataset. Given the left-most molecule, we interpolate in latent space along the direction which maximizes its QED property.}
    \label{fig:prop_zinc}
\end{figure}

\begin{figure}[!htp]
    \centering
    \includegraphics[scale=0.4]{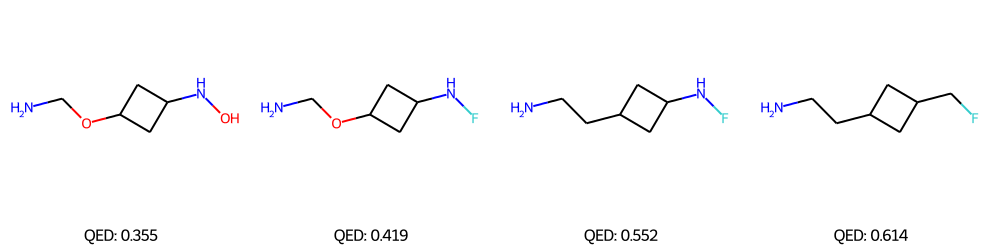}
    \caption{Example of chemical property optimization on the QM9 dataset. Given the left-most molecule, we interpolate in latent space along the direction which maximizes its QED property.}
    \label{fig:prop_qm9}
\end{figure}

\begin{table}[!htp]
    \caption{Performance in terms of the best QED scores (baselines are taken from \citet{graphdf}).}
    \label{tab:add_results3}
    \centering
    %\resizebox{\textwidth}{!}{
    \begin{tabular}{lccc}
        \toprule
        Method & 1st & 2nd & 3rd     \\
        \midrule
        ZINC (dataset)    &  0.948    &  0.948    &  0.948       \\
        \midrule
        JTVAE & 0.925 & 0.911& 0.910    \\
        GCPN  & 0.948 & 0.947& 0.945        \\
        MRNN  & 0.844 & 0.799& 0.736     \\
        GraphAF  & 0.948 & 0.948& 0.947     \\
        GraphDF  & 0.948 & 0.948& 0.948     \\
        MoFlow & 0.948    & 0.948 &0.948     \\
        \midrule
        \atMod~ (2D-EGNN)  & 0.948 & 0.941 & 0.937  \\
        \atMod~ (3D-EGNN)  & 0.948 & 0.937 & 0.931    \\
        \atMod~ (JT-2D-EGNN) & 0.947 & 0.941  & 0.939   \\
        \atMod~ (JT-3D-EGNN)  & 0.948 &  0.948 & 0.945   \\
        \bottomrule
    \end{tabular}
    %}
\end{table}

\autoref{fig:prop_zinc} and \autoref{fig:prop_qm9} show examples of the molecules decoded from the learned latent space using this procedure, starting with molecules having a low QED score. Note that  the number of valid molecules decoded back varies on the query molecule. We report the discovered novel molecules sorted by their QED scores in  \autoref{tab:add_results3}. Clearly, \atMod~ is able to find novel molecules with high QED scores. 

\subsection{Ablation Studies}
We also performed ablation experiments to gain further insights about \atMod, as we describe next.

\paragraph{E(3)-equivariant versus not equivariant.}
Molecules exhibit translational and rotational symmetries, so we conducted an ablation study to quantify the effect of incorporating these symmetries in our model. We compare the results obtained using an EGNN with a non-equivariant graph convolutional network (GCN). For our purpose, we used a 3-layer GCN with layer sizes 64-32-32. The validity scores in \autoref{tab:ablation_equiv} provide strong evidence in favor of modeling the symmetries explicitly in the proposed Modular Flows.
\begin{table}[!h]
    \caption{Random generation performance on ZINC250K and QM9 dataset with E(3)-EGNN vs GCN.}
    \label{tab:ablation_equiv}
    \centering
    \resizebox{0.8\textwidth}{!}{
    \begin{tabular}{llcccc}
        \toprule
        Dataset & Method  &  Validity \%  & Uniqueness \% & Novelty \% \\
        \midrule
        %& \atMod (2D-EGNN) & ($93.8 \pm 0.1$) & $93.8 \pm 0.1$ & 99.9 & 100   \\
        \multirow{2}{4em}{ZINC250K} & \atMod ~(3D-EGNN) &  $95.4 \pm 1.2$ & 99.7 & 100  \\
        & \atMod ~(GCN) &  $90.3 \pm 1.9 $ & 99.7 & 100   \\
       \multirow{2}{4em}{QM9}  & \atMod ~(3D-EGNN) &  $98.3 \pm 0.7$ & 99.1 & 100   \\
        & \atMod ~(GCN) &  $93.3 \pm 0.5$ & 98.8  & 100   \\
        \bottomrule
    \end{tabular}}
\end{table}
\paragraph{2D versus 3D.}
Finally, we study whether including information about the 3D coordinates improves the model. Note that the EGNN-coupled differential function obtains either the 2D or 3D positions as polar coordinates, where the 3D positions have an extra degree of freedom. \autoref{tab:ablation_2d3d} shows that transitioning from 2D to 3D improves the mean validity score. 
\begin{table}[!h]
    \caption{Random generation on ZINC250K and QM9 dataset with 2D versus 3D features.}
    \label{tab:ablation_2d3d}
    \centering
    \resizebox{0.8\textwidth}{!}{
    \begin{tabular}{llcccc}
        \toprule
        Dataset & Method  &  Validity \%  & Uniqueness \% & Novelty \% \\
        \midrule
        %& \atMod (2D-EGNN) & ($93.8 \pm 0.1$) & $93.8 \pm 0.1$ & 99.9 & 100   \\
        \multirow{2}{4em}{ZINC250K} & \atMod ~(3D-EGNN) &  $95.4 \pm 1.2$ & 99.7 & 100  \\
        & \atMod ~(2D-EGNN) &  $94.8 \pm 1.0$ & 99.4 & 100   \\
       \multirow{2}{4em}{QM9}  & \atMod ~(3D-EGNN) &  $98.3 \pm 0.7$ & 99.1 & 100   \\
        & \atMod ~(2D-EGNN) & $96.2 \pm 1.7$ & 99.5 & 100   \\
        \bottomrule
    \end{tabular}}
\end{table}
%\input{Discussion}

%\section{Discussion}

%We have explored the efficacy and elegance of graph PDEs in molecular generation. However, this area still remains to be unexplored where PDEs provide a flexible and compatible representation of complex objects, which paves the way for 

% - modular diffusion (just mention SDE)
% - invertibility pathology of normalizing flows: inversion of softmax or argmax
% - molecular 3D point clouds (a 3D grid expansion like climate pde etc)
% - Adjoints, checkpoints
% - conditional, control, targeted
% - probabilistic treatment of continuous->discrete flows
% - multi-layer vs single-layer EGNN differential
\section{Conclusion}
We proposed \atMod, a new generative flow model where multiple flows interact locally according to a coupled ODE, resulting in accurate modeling of graph densities and high quality molecular generation without any validity checks or correction. Interesting avenues open up, including the design of (a) more nuanced mappings between discrete and continuous spaces, and (b) extensions of modular flows to (semi-)supervised settings.

\section{Acknowledgments}
The calculations were performed using resources within the Aalto University Science-IT project. This work has been supported by the Academy of Finland under the {\em HEALED} project (grant 13342077).

% The use of modular coupled ODE results in simple but high-capacity model with empirically more accurate modelling of graph structure densities. Experimental results show that ~\atMod~ outperforms all previous state-of-the-art baselines on the standard tasks. Our model achieves very high validity, and is close to solving the validity problem for good. 

% While the simplicity and expressiveness of PDE-based graph modelling is pleasing, there are further open problems. From wider perspective, the mappings between discrete and continuous flow spaces warrants further research \citep{papamakarios2021normalizing}. Expanding modular flows to supervised learning by simultaneously modelling property surfaces is an interesting future direction, as is using flow models in reinforcement learning settings.
% %Speficically, the non-invertibility of the commonly used softmax and argmax sampling steps requires further work.

%\newpage
\small{\bibliography{ref}}
\small {\bibliographystyle{plainnat}}
%\small {\bibliographystyle{unsrt}}

\newpage

\appendix
\newpage
\section{Appendix}

\subsection{Derivation of modular adjoint}
\label{sec:adjoint}
We present a standard adjoint gradient derivation \citep{bradley2019}, and show that the adjoint of a graph neighborhood differential is sparse.

For completeness, we define an ODE system
\begin{align}
    \dz(t) &= f(\z,t,\bt) \\
    \z(t) &= \z_0 + \int_0^t f(\z,t,\bt) d\tau,
\end{align}
where $\z \in \R^D$ is a state vector, $\dz \in \R^D$ is the state time differential defined by the vector field function $f$ and parameterised by $\bt$. The starting state is $\z_0$, and $t,\tau \in \R_+$ are time variables. Our goal is to solve a constrained problem
\begin{align}
    \min_\bt \qquad & G(\bt) = \int_0^T g(\z, t, \bt) dt \\
     s.t. \qquad &  \dz - f(\z,t,\bt) = 0, \qquad \forall t \in [0,T] \\
      \qquad & \z(0) - \z_0 = 0,
\end{align}
where $G$ is the total loss that consists of instant loss functionals $g$. We desire to compute the gradients of the system $\nabla_\bt G$.

We optimise the constrained problem by solving the Lagrangian 
%\color{blue} in order to avoid confusion due to overload with time $T$, changing transpose operator (used in Lagrangian) below to $\top$ instead \color{black}
\begin{align}
    \L(\bt,\bl,\bmu) &= G(\bt) + \int_0^T \bl(t)^{\top} (\dz - f(\z,t,\bt)) dt + \bmu^{\top} (\z(0) - \z_0) \\
    &= \int_0^T \Big[g(\z,t,\bt) + \bl(t)^{\top} (\dz - f(\z,t,\bt))\Big] dt + \bmu^{\top} (\z(0) - \z_0)~.
\end{align}
The constraints are satisfied by the ODE definition. Hence, $\nabla_\bt \L = \nabla_\bt G$, and we can set values $\bt$ and $\bmu$ freely. We use a shorthand notation  $\frac{\partial a}{\partial b} = a_b$, and omit parameters from the functions for notational simplicity. Applying the chain rule, we note that the  gradient becomes
\begin{align}
    \nabla_\bt \L &= \nabla_\bt G = \int_0^T \Big[\color{black}g_\z \z_\bt + g_\bt + \bl^{\top} \dz_\bt - \bl^{\top} f_\z \z_\bt - \bl^{\top} f_\bt \Big]\color{black} dt~,
\end{align}
where the $\bmu$ term drops out since it does not depend on parameters $\bt$. We apply integration by parts to swap the differentials in term $\bl^{\top} \dz_\bt$, resulting in 
\begin{align}
    %\int_0^T \bl^{\top} \dz_\bt dt &= \bl^{\top} \dz_\bt|_{t=T} - \bl^{\top} \dz_\bt|_{t=0} - \int_0^T \dot{\bl}^{\top} \z_\bt dt~.
     \int_0^T \bl^{\top} \dz_\bt dt &= \bl^{\top} \z_\bt|_{t=T} - \bl^{\top} \z_\bt|_{t=0} - \int_0^T \dot{\bl}^{\top} \z_\bt dt~.
\end{align}

%\color{blue} Shouldn't the above equation be

%\begin{align}
%    \int_0^T \bl^{\top} \dz_\bt dt &= \bl^{\top} \z_\bt|_{t=T} - \bl^{\top} \z_\bt|_{t=0} - \int_0^T \dot{\bl}^{\top} \z_\bt dt~.
%\end{align}

\color{black}

Substituting this into previous equation and rearranging the terms results in
\begin{align}
    %\nabla_\bt \L &= \int_0^T \underbrace{(g_\z - \bl^{\top} f_\z - \dot{\bl}^{\top}) \z_\bt}_{0, \text{ if } \dot{\bl}^{\top} = g_\z - \bl^{\top} f_\z} dt + \int_0^T g_\bt - \bl^{\top} f_\bt dt + \underbrace{\bl^{\top} \dz_\bt|_{t=T}}_{0, \text{ if } \bl(T) = 0} - \underbrace{\bl^{\top} \dz_\bt|_{t=0}}_{0}.
    \nabla_\bt \L &= \int_0^T \underbrace{(g_\z - \bl^{\top} f_\z - \dot{\bl}^{\top}) \z_\bt}_{0, \text{ if } \dot{\bl}^{\top} = g_\z - \bl^{\top} f_\z} dt + \int_0^T (g_\bt - \bl^{\top} f_\bt) dt + \underbrace{\bl^{\top} \z_\bt|_{t=T}}_{0, \text{ if } \bl(T) = \0} - \underbrace{\bl^{\top} \z_\bt|_{t=0}}_{0}.
\end{align}

%\color{blue}
%The above equation should probably be
%\begin{align}
%    \nabla_\bt \L &= \int_0^T \underbrace{(g_\z - \bl^{\top} f_\z - \dot{\bl}^{\top}) \z_\bt}_{0, \text{ if } \dot{\bl}^{\top} = g_\z - \bl^{\top} f_\z} dt + \int_0^T (g_\bt - \bl^{\top} f_\bt) dt + \underbrace{\bl^{\top} \z_\bt|_{t=T}}_{0, \text{ if } \bl(T) = \0} - \underbrace{\bl^{\top} \z_\bt|_{t=0}}_{0}.
%\end{align}
\color{black}

The last term is removed since $\z(0)$ not depend on $\bt$ as a constant, and thus $\z_\bt(0)=0$. The difficult term in the equation is $\z_\bt$. We remove it by choosing 
\begin{align}
    \dot{\bl}^{\top} &= g_\z - \bl^{\top} f_\z.
\end{align}
Finally, we choose $\bl(T)=0$ which also removes the second-to-last term. The choices lead to a final term
\begin{align}
    \nabla_\bt G = \nabla_\bt \L &= \int_0^T (\color{black}g_\bt - \bl^{\top} f_\bt)\color{black} dt \\
    s.t. \quad & \dot{\bl}^{\top} = g_\z - \bl^{\top} f_\z \label{eq:dlambda} \\
    & \bl(T) = 0.
\end{align}
In the derivation the adjoint $\bl(t) = \frac{\partial \L}{\partial \z(t)} \in \R^D$ represents the change of loss with respect to instant states, and is another ODE system that runs backwards from $\bl(T) = 0$ until $\bl(0)$. The final gradient $\nabla_\bt \L$ counts all adjoints within $[0,T]$ multiplied by the `immediate' partial derivatives $f_\bt$. The final gradient also takes into account the instant loss parameter derivatives. For simple MSE curve fitting, the instant loss has no parameters. The adjoint depends on the instant loss state derivatives $g_\z$. These are often only available for observations $\y_j$ at observed timepoints $t_j$. This can be represented by having a convenient loss
\begin{align}
    g(\z,t,\bt) &= \delta(t=t_j) \tilde{g}(\z,\y_j,t,\bt),
\end{align}
and now the term $g_\z$ induces discontinuous jumps at observations. This does not pose problems in practice, since we can integrate the ODE in continuous segments between the observation instants.

The sparsity of the adjoint evolution is evident from \autoref{eq:dlambda}, where the $\dot{\bl}_i$ is an inner product between $\bl$ and one column of $\frac{\partial f}{\partial \z}$, which is invariant to non-neighbors. This gives the result
\begin{align}
    \frac{d\bl_i}{dt} &= - \bl(t)^{\top} \frac{\partial f\big(t, \z_i(t), \z_{\N_i}(t),\mathbf{x}_{i}, \mathbf{x}_{\N_i}\big) }{\partial \z} = - \sum_{j \in \N_i \cup \{i\}} \bl_j(t)^{\top} \frac{\partial f\big(t, \z_i(t), \z_{\N_i}(t),\mathbf{x}_{i}, \mathbf{x}_{\N_i}\big) }{\partial \z_j}.
\end{align}

\subsection{Tree Decomposition}
\label{sec:tree}
\begin{figure}[!htp]
\centering
\begin{subfigure}{.5\textwidth}
  \centering
  \includegraphics[width=.95\linewidth]{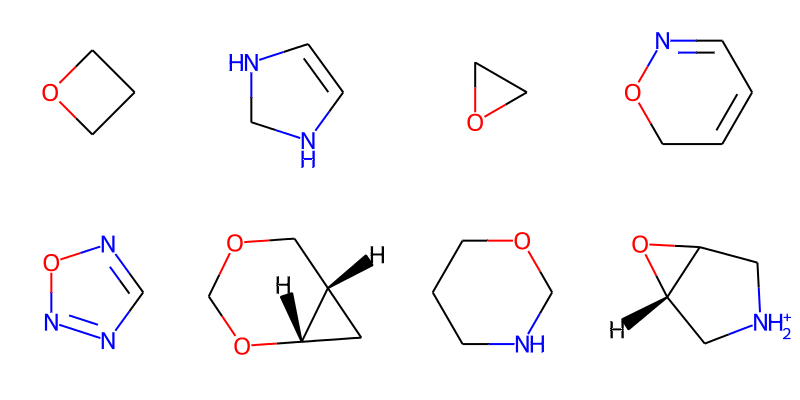}
  \caption{QM9 Dataset}
  \label{fig:ring_qm9}
\end{subfigure}%
\begin{subfigure}{.5\textwidth}
  \centering
  \includegraphics[width=.95\linewidth]{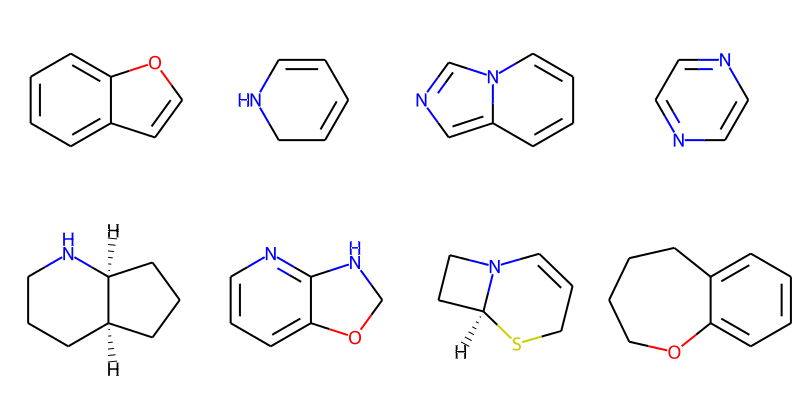}
  \caption{ZINC250K Dataset}
  \label{fig:ring_zinc}
  \end{subfigure}
\caption{Examples of frequently occurring ring substructures}
\label{fig:ring_mol}
\end{figure}

For tree decomposition of the molecules, we followed closely the procedure described in \citet{jin2018junction}. The rings as well as the nodes corresponding to each ring substructure were extracted using RDKit's functions, \texttt{GetRingInfo} and \texttt{GetSymmSSSR}. We restricted our vocabulary to the unique ring substructures in the molecules. The vocabulary of clusters follows a skewed distribution over the frequency of appearance within the dataset. In particular, only a subset ($\sim$ 30) of ring substructures (labels) appear with high frequency in molecules within the vocabulary. Therefore, we simplify the vocabulary by only representing the 30 commonly occurring substructures of $\mathcal{A}_{tree}$. In \autoref{fig:ring_mol}, we show some examples of these ring substructures for the two datasets. 

\subsection{Equivariant Graph Neural Networks }
\label{sec:egnn}
Equivariant Graph Neural Networks (EGNN) \citep{egnn} are E(3)-equivariant with respect to an input set of points. The E(3) equivariance accounts for translation, rotation, and reflection symmetries, and can be extended to E($n$) group equivariance. The inherent dynamics governing the EGNN can be described, for each layer $l$, as follows. Here, $\mathbf{h}_{i}^{l}$ and $\mathbf{x}_{i}^{l}$ pertain, respectively, to the embedding for the node $i$ and that for its coordinates; and $a_{ij}$ abstracts the information about the edge between nodes $i$ and $j$.  

 %For input the set of node embeddings $\mathbf{h}^{l} = \{ \mathbf{h}_{0}^{l}, \mathbf{h}_{1}^{l},...,\mathbf{h}_{M-1}^{l}\}$ and coordinate embeddings $\mathbf{x}^{l} = \{ \mathbf{x}_{0}^{l}, \mathbf{x}_{1}^{l},...,\mathbf{x}_{M-1}^{l}\}$ and edge
%information E = ($a_{ij}$ ) and outputs a transformation on $\mathbf{h}^{l+1}$ and $\mathbf{x}^{l+1}$ as:
\begin{align} 
\mathbf{m}_{i j} &=\phi_{e}\left(\mathbf{h}_{i}^{l}, \mathbf{h}_{j}^{l},\left\|\mathbf{x}_{i}^{l}-\mathbf{x}_{j}^{l}\right\|^{2}, a_{i j}\right) \nonumber \\
\mathbf{x}_{i}^{l+1} &=\mathbf{x}_{i}^{l}+C \sum_{j \neq i}\left(\mathbf{x}_{i}^{l}-\mathbf{x}_{j}^{l}\right) \phi_{x}\left(\mathbf{m}_{i j}\right) \label{eq:egnn_eq} \nonumber\\
\mathbf{m}_{i} &=\sum_{j \neq i} \mathbf{m}_{i j} \nonumber \\
\mathbf{h}_{i}^{l+1} &=\phi_{h}\left(\mathbf{h}_{i}^{l}, \mathbf{m}_{i}\right) \nonumber
\end{align}

Initially, messages $\mathbf{m}_{i j}$ are computed between the neighboring nodes via $\phi_{e}$.  Subsequently, the coordinates of each node $i$ are updated via a weighted sum of relative position vectors $\{(\mathbf{x}_{i}-\mathbf{x}_{j}) : j \neq i\}$ with the aid of $\phi_{x}$.  Finally, the node embeddings are updated based on the aggregated messages $\mathbf{m}_{i}$ via $\phi_{h}$. The aggregated message can be computed based on only the neighbors of a node by simply replacing the sum over $j \neq i$ with a sum over $j \in \mathcal{N}_i$ in these equations.

%Here, the edge operation $\phi_{e}$ also take  input the relative squared distance between two coordinates $\left\|\mathbf{x}_{i}^{l}-\mathbf{x}_{j}^{l}\right\|^{2}$, with node embeddings $\mathbf{h}_{i}^{l}, \mathbf{h}_{j}^{l}$ and edge attribute $a_{i j}$. Subsequently, the position of each node $\mathbf{x}_{i}^{l}$ is updated via a vector field in a radial direction. This means that the position is updated by the weighted sum of all relative differences $(\mathbf{x}_{i}-\mathbf{x}_{j})_{\forall j}$ by function $\phi_{x}$. C is chosen to be $1/(M - 1)$, which divides the sum by its number of elements. Eq. 18 is the aggregation step, which can be limited to the neighborhood $j \in \mathcal{N}(i)$, and Eq. 19 performs node operation $\phi_{h}$. In case, when dealing with static coordinates $\mathbf{x}_{i}$ this becomes E(n) invariant.  

\subsection{Connection to Temporal Graph Networks}
\label{sec:temporal_to_egnn}

Temporal Graph Networks \citep{temporal,PINT2022}  
are state-of-the-art neural models for embedding dynamic graphs. A prominent class of these models consists of a combination of (recurrent) memory modules and graph-based operators, and rely on message passing for updating the embeddings based on node-wise or edge-wise {\em events}. 

Specifically, adopting the notation from \citet{temporal}, an interaction $\mathbf{e}_{i j}(t)$ between any two nodes $i$ and $j$ at time $t$ triggers an edge-wise event leading to the following steps. First, a message $\mathbf{m}_{ij}'(t)$ is computed based on the memory $\mathbf{s}_{i}\left(t^{-}\right)$ and $\mathbf{s}_{j}\left(t^{-}\right)$ of the two nodes just before time $t$ via a learnable function $\operatorname{msg}$ (such as multilayer perceptron). For each node $i$, the messages thus accrued over a small period due to interactions of with neighbors $j$ are combined (via $\operatorname{agg}$) into an aggregate message $\overline{\mathbf{m}}_{i}'(t)$. This message, in turn, is used to update the memory of $i$ to $\mathbf{s}_{i}(t)$ via $\operatorname{mem}$ (implemented e.g., as a recurrent neural network). Finally, the node embedding of $i$ is revised based on its memory $\mathbf{s}_{i}(t)$, interaction $\mathbf{e}_{i j}(t)$ and memory  $\mathbf{s}_{j}(t)$ of each neighbor $j \in \mathcal{N}_i$, as well as any additional node-wise events $\v_i(t)$ involving $i$ or any node in $\mathcal{N}_i$. 

\begin{table}[!t]
\centering
    \caption{\atMod~ as a temporal graph network (TGN). Adopting notation for TGNs from \cite{temporal} $v_i$ is a node-wise event on $i$; $e_{ij}$ denotes an (asymmetric) interaction between $i$ and $j$; $\mathbf{s}_{i}$ is the memory of node $i$; and $t$ and $t^-$ denote time with $t^-$ being the time of last interaction before $t$,  e.g., $\mathbf{s}_{i}(t^-)$ is the memory of $i$ just before time $t$; and $\operatorname{msg}$ and $\operatorname{agg}$ are learnable functions (e.g., MLP) to compute, respectively, the individual and the aggregate messages. For \atMod, we use   $\mathbf{r}_{ij}$ to denote the spatial distance  $\mathbf{x}_{i} - \mathbf{x}_{j}$, and $a_{i j}$ to denote the attributes of the edge between $i$ and $j$. The functions $\phi_e$, $\phi_x$, and $\phi_h$ are as defined in \cite{egnn}, and summarized in \ref{sec:egnn}.   \label{tab:egnn_tgn_mod}}
    \centering
    \resizebox{\textwidth}{!}{
    \begin{tabular}{lcc}
        \toprule
        Method  & TGN layer &  \atMod   \\
        \midrule
        
        \multirow{2}{4em}{Edge}  & $\mathbf{m}_{i j}'(t)=\operatorname{msg}\left(\mathbf{s}_{i}\left(t^{-}\right), \mathbf{s}_{j}\left(t^{-}\right),  \Delta t, \mathbf{e}_{ij}(t)\right)  $ & $\mathbf{m}_{i j}(t) =\phi_{e}\left(\mathbf{z}_{i}(t), \mathbf{z}_{j}(t),\left\|\mathbf{r}_{ij}(t)\right\|^{2}, a_{i j}\right)$   \\
        
         & $\overline{\mathbf{m}}_{i}'(t)=\operatorname{agg}\left(\{\mathbf{m}_{i j}'\left(t\right) | j \in  \N_i \}\right) $ & $\mathbf{m}_{i}(t) = \sum_{j\in \mathcal{N}(i)} \mathbf{m}_{ij} $   \\
         & & $\hat{\mathbf{m}}_{ij}(t) = \mathbf{r}_{ij}(t)\cdot \phi_{x}\left(\mathbf{m}_{i j}(t)\right) $ \\
         &  & $\hat{\mathbf{m}}_{i}(t) = C \sum_{j\in \mathcal{N}(i)} \hat{\mathbf{m}}_{ij}(t) $   \\
         \midrule
        Memory state  & $\mathbf{s}_{i}(t)=\operatorname{mem}\left(\overline{\mathbf{m}}_{i}'(t), \mathbf{s}_{i}\left(t^{-}\right)\right)$  &$\mathbf{x}_{i}(t+1) = \mathbf{x}_{i}(t)+ \hat{\mathbf{m}}_{i}(t)$   \\
        \midrule
        Node  & $\mathbf{z}_{i}'(t)=\sum_{j \in \N_i} h\left(\mathbf{s}_{i}(t), \mathbf{s}_{j}(t), \mathbf{e}_{i j}(t), \mathbf{v}_{i}(t), \mathbf{v}_{j}(t)\right)$ & $\mathbf{z}_{i}(t+1) =\phi_{h}\left(\mathbf{z}_{i}(t), \mathbf{m}_{i}(t)\right)$  \\
        \bottomrule
    \end{tabular}}
\end{table}

%The role of the memory state $\mathbf{s}_{i}$ in TGN can be played by coordinate embeddings $\mathbf{x}_{i}$ in our method and messages can be formulated with respect to both in the networks. It is worthy to note that TGN also incorporates neighbouring node attributes, memory states to update the node embedding $\mathbf{z}_{i}'$ which is the similar to ours where the aggregated message over neighbouring nodes is used to update the node embedding $\mathbf{z}_{i}$, where ($\phi_{h}$) is a  learnable function. From the above similarities, we can see that the our method can be casted as a variant of TGN, which is also equivariant as compared to TGN. 
It turns out (see \autoref {tab:egnn_tgn_mod}) that \atMod~ can be viewed as an equivariant message passing temporal graph network. Interestingly, the coordinate embedding $\mathbf{x}_{i}$ plays the role of the memory $\mathbf{s}_{i}$.

\subsection{Implementation Details}
\label{sec:implement}
We implemented the proposed models in PyTorch \citep{paszke2019pytorch}.\footnote{We make the code available at \url{https://github.com/yogeshverma1998/Modular-Flows-Neurips-2022}.} We used a single layer for EGNN with embedding dimension 32 and aggregated information for each node from only its immediate neighbors. For geometric (spatial) information, we worked with the polar coordinates (2D) or the spherical polar coordinates (3D). We solved the ODE system with the Dormand–Prince adaptive step size scheme (i.e., the \texttt{dopri5} solver). The number of function evaluations lay roughly between $70$ and $100$. The models were trained for 50-100 epochs with the Adam \citep{kingma2014adam} optimizer. 

\paragraph{Time comparisons.} We found the training time of \atMod~ to be slightly worse than one-shot discrete flow models that characterize the whole system using a single flow (recall that, in contrast, \atMod~ associates an ODE with each node). However, \atMod~ is faster to train than the auto-regressive methods. 

Note that computation is a crucial aspect of generative modeling for application domains with a huge search space, as is true for the molecules. We report the computational effort (excluding the time for preprocessing) for generating 10000 molecules averaged across 5 independent runs in \autoref{tab:time_1}.  Notably, largely by virtue of being one-shot, \atMod~ is able to generate significantly faster than the auto-regressive models such as GraphAF and GraphDF. \atMod~ also owes this speedup, in part, to obviate the need for multiple decoding (unlike, e.g., JT-VAE) as well as any validity checks.

\begin{table}[!h]
    \caption{Generation time (in seconds/molecule) on QM9 and ZINC250K.}
    \label{tab:time_1}
    \centering
    %\resizebox{\textwidth}{!}{
    \begin{tabular}{lcc}
        \toprule
        Method & ZINC250K & QM9   \\
        \midrule
        GraphEBM & 1.12 $\pm$ 0.34 &0.53 $\pm$ 0.16 \\
        GVAE      & 0.86 $\pm$ 0.12   & 0.46 $\pm$ 0.07   \\
        GraphAF  &0.93 $\pm$ 0.14  & 0.56 $\pm$ 0.12  \\
        GraphDF  & 3.12 $\pm$ 0.56 & 1.92 $\pm$ 0.42  \\
        MoFlow & 0.71 $\pm$ 0.14   &  0.31 $\pm$ 0.04  \\
        \midrule
        \atMod~ (2D-EGNN)  &0.46 $\pm$ 0.09 & 0.16 $\pm$ 0.04  \\
        \atMod~ (3D-EGNN)  & 0.55 $\pm$ 0.13 & 0.24 $\pm$ 0.06  \\
        \atMod~ (JT-2D-EGNN) &0.53 $\pm$ 0.07 & 0.21 $\pm$ 0.07  \\
        \atMod~ (JT-3D-EGNN)  &0.62 $\pm$ 0.11 & 0.28 $\pm$ 0.09   \\
        \bottomrule
    \end{tabular}
    %}
\end{table}

\subsection{Additional Evaluation Metrics}
\label{sec:moses}

We invoked additional metrics, namely the MOSES metrics \citep{moses}, to compare the different models in terms of their ability to generate molecules. These metrics, described below, access the overall quality of the generated molecules. 

\begin{itemize}
    \item \textbf{FCD}: {\em Fr\'{e}chet Chemnet Distance} (FCD) \citep{preuer2018frechet} is a general purpose metric that measures diversity of the generated molecules, as well as the extent of their chemical and biological property alignment with a reference set of real molecules. Specifically, the last layer activations of ChemNet are used for this purpose. Lower is better. 
    \item \textbf{Frag}: {\em Fragment similarity} (Frag), measures the cosine distance between the fragment frequencies of the generated molecules and a set of reference molecules. Higher is better.
    \item \textbf{SNN}: {\em Nearest Neighbor Similarity} (SNN) quantifies how close the generated molecules are to the true molecule manifold. Specifically, it computes the average similarity of a generated molecule to the nearest molecule from the reference set. Higher is better.
    \item \textbf{IntDiv}: As the name suggests, {\em Internal Diversity} (IntDiv) accounts for diversity by computing the average pairwise similarity of the generated molecules. Higher is better. 
    %\item To measure the diversity, chemical and biological similarity of generated molecules, we used  Fréchet ChemNet Distance (FCD)\ which uses last layer activation's of ChemNet to measure the differences.
\end{itemize}
For our purpose, we evaluated these metrics with QM9 and ZINC250K as the reference sets. As shown in \autoref{tab:add_results1} and \autoref{tab:add_results2}, \atMod~ achieves better performance results across all metrics. Notably, \atMod~ registers lower FCD and higher IntDiv scores compared to other methods, suggesting that \atMod~  is able to generate diverse set of molecules similar to those present in the real datasets.

\begin{table}[!htp]
    \caption{Evaluation of performance on MOSES metrics on generative models on QM9 dataset. FCD is lower the better, Frag, SNN, and IntDiv higher the better.  }
    \label{tab:add_results1}
    \centering
    %\resizebox{\textwidth}{!}{
    \begin{tabular}{lcccc}
        \toprule
        Method & FCD ($\downarrow$) & Frag ($\uparrow$)& SNN ($\uparrow$) & IntDiv ($\uparrow$)    \\
        \midrule
        GVAE    &  0.513    &  0.821    &  0.582     & 0.822         \\
        GraphEBM & 0.551 & 0.831& 0.547 &  0.831   \\
        GraphAF  & 0.732 & 0.863& 0.565 & 0.823      \\
        GraphDF  & 0.683 & 0.892&0.562  &  0.839     \\
        MoFlow & 0.496  & 0.840 &0.502 &  0.852      \\
        \midrule
        \atMod~ (2D-EGNN)  & 0.432 & 0.928 & 0.608 & 0.875   \\
        \atMod~ (3D-EGNN)  & 0.478 & 0.934 & 0.613 & 0.885   \\
        \atMod~ (JT-2D-EGNN) & 0.421& 0.921& 0.595 & 0.867   \\
        \atMod~ (JT-3D-EGNN)  & 0.401 & 0.939 & 0.624 & 0.889  \\
        \bottomrule
    \end{tabular}
    %}
\end{table}
\begin{table}[!htp]
    \caption{Evaluation of performance on MOSES metrics on generative models on ZINC250K dataset. FCD is lower the better, Frag, SNN, and IntDiv  higher the better.  }
    \label{tab:add_results2}
    \centering
    %\resizebox{\textwidth}{!}{
    \begin{tabular}{lcccc}
        \toprule
        Method & FCD ($\downarrow$) & Frag ($\uparrow$)& SNN ($\uparrow$) & IntDiv ($\uparrow$)    \\
        \midrule
        JTVAE  &    0.512    &    0.890   &   0.5477    & 0.855     \\
        GVAE    &  0.571    &  0.871    &  0.532     & 0.852        \\
        GraphEBM & 0.613 & 0.843& 0.487 &  0.821   \\
        GraphAF  & 0.524 & 0.803& 0.465 & 0.855       \\
        GraphDF  & 0.658 & 0.869& 0.515 & 0.829    \\
        MoFlow & 0.597    & 0.851 &0.452 &  0.832    \\
        \midrule
        \atMod~ (2D-EGNN)  & 0.495 & 0.891 & 0.570 & 0.863 \\
        \atMod~ (3D-EGNN)  & 0.512& 0.905 & 0.584 & 0.869  \\
        \atMod~ (JT-2D-EGNN) & 0.501& 0.915 & 0.563 & 0.857 \\
        \atMod~ (JT-3D-EGNN)  & 0.523 & 0.929 & 0.594 & 0.879 \\
        \bottomrule
    \end{tabular}
    %}
\end{table}

\newpage

 We also evaluated the generated structures via distributions of their important properties. Specifically, we obtained kernel density estimates of these distributions to aid in visualization. We consider the following standard properties. 

\begin{itemize}
    \item \textbf{Weight}: sum of the individual atomic weights of a molecule. The weight provides insight into the bias of the generated molecules toward lighter or heavier molecules.
    \item \textbf{LogP}: ratio of concentration in octanol-phase to the aqueous phase, also known as the octanol-water partition coefficient. It is computed via the Crippen \citep{logp} estimation.
    \item \textbf{Synthetic Accessibility (SA)}: an estimate for the synthesizability of a given molecule. It is calculated based on contributions of the molecule fragments \cite{sas}. \iffalse and does not adequately assess up-to-date chemical structures \fi 
    \item \textbf{Quantitative Estimation of Drug-likeness (QED)}: describes the likeliness of a molecule as a viable candidate for a drug. It ranges between [0,1] and captures the abstract notion of aesthetics in medicinal chemistry \citep{qed}.
\end{itemize}

\begin{figure}[!htp]
    \centering
    \includegraphics[scale=0.23]{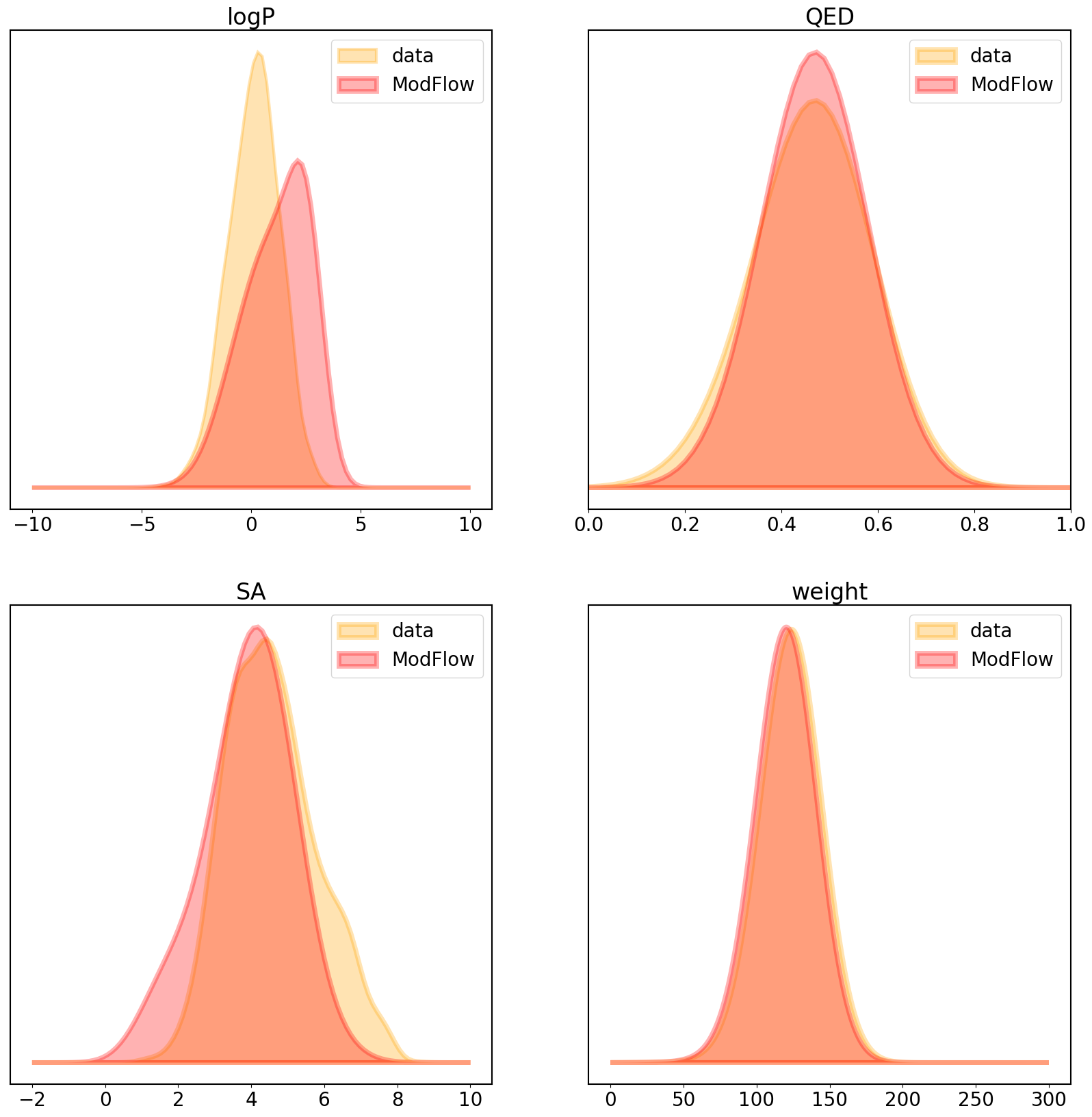}
    \caption{\textbf{(QM9)} Distributions of the chemical properties.}
    \label{fig:kde_QM9}
\end{figure}

\begin{figure}[!htp]
    \centering
    \includegraphics[scale=0.23]{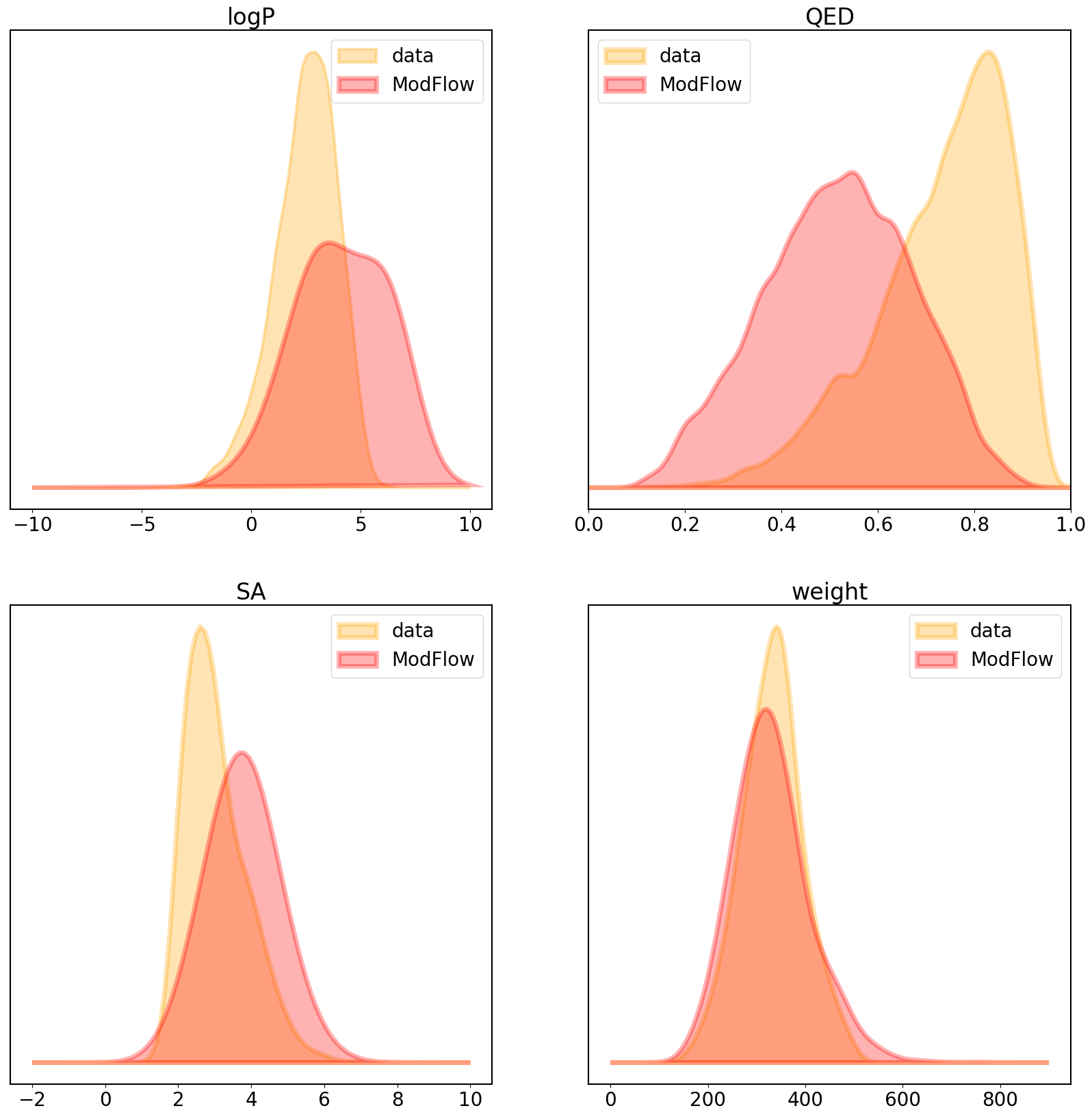}
    \caption{\textbf{(ZINC250K)} Distributions of the chemical properties.}
    \label{fig:kde_ZINC}
\end{figure}

\autoref{fig:kde_QM9} and \autoref{fig:kde_ZINC}
show that barring some dispersion in QED and logP (especially on Zinc250K), the property distributions of the molecules generated by \atMod~ generally match the corresponding distributions on the reference datasets quite closely. These results demonstrate the effectiveness of \atMod~ in generating molecules that have properties similar to  the molecules in the reference set.

\newpage

\newpage 

\subsection{Additional examples of molecules generated by \atMod~.} \label{sec:GenMols}
\begin{figure}[!h]
    \centering
    \includegraphics[scale=0.4, trim={0 0 0 2cm}, clip]{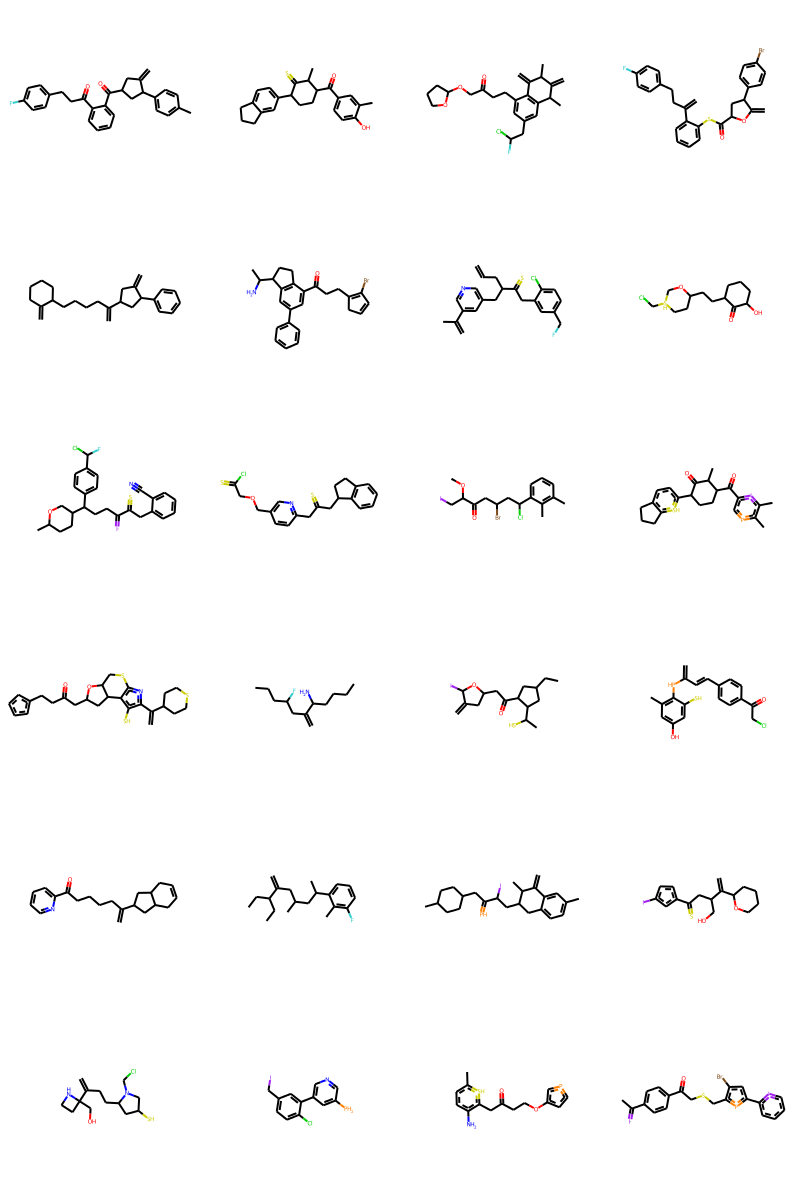}
   % \caption{Grid of randomly generated molecules from our model}
    \label{fig:gen_mol}
\end{figure}

\end{document}